\documentclass[twoside]{article}
\usepackage{ecj,palatino,epsfig,latexsym}
\usepackage[square, numbers]{natbib}

\usepackage[english]{babel}
\usepackage{tabularx}
\usepackage[table]{}

\usepackage[cmex10]{amsmath}
\usepackage{epsfig}
\usepackage{fixltx2e}
\usepackage{amssymb}
\usepackage{multirow}

\usepackage{graphicx}
\usepackage{amsmath,amssymb}
\usepackage[boxed]{algorithm2e}

\begin{document}

\ecjHeader{x}{x}{xxx-xxx}{200X}{General Subpopulation Framework}{Vargas et al.}
\title{\bf General Subpopulation Framework and Taming the Conflict Inside Populations}

\author{\name{\bf Danilo Vasconcellos Vargas} \hfill \addr{vargas@cig.ees.kyushu-u.ac.jp}\\ 
        \addr{Graduate School of Information Science and Electrical Engineering, Kyushu University, Fukuoka, 819-0395, Japan}
\AND
	\name{\bf Junichi Murata} \hfill \addr{murata@cig.ees.kyushu-u.ac.jp}\\
	\addr{Faculty of Information Science and Electrical Engineering, Kyushu University, Fukuoka, 819-0395, Japan}
\AND
	\name{\bf Hirotaka Takano} \hfill \addr{takano@cig.ees.kyushu-u.ac.jp}\\
	\addr{Faculty of Information Science and Electrical Engineering, Kyushu University, Fukuoka, 819-0395, Japan}
\AND
       \name{\bf Alexandre Cl\'{a}udio Botazzo Delbem} \hfill \addr{acbd@icmc.usp.br}\\
        \addr{Institute of Mathematics and Computer Science, University of S\~{a}o Paulo, 
        S\~{a}o Carlos, 13566-590, Brazil}
}


\maketitle

\begin{abstract}
Structured evolutionary algorithms have been investigated for some time. 
However, they have been under-explored specially in the field of multi-objective optimization. 
Despite their good results, the use of complex dynamics and structures make their understanding and adoption rate low.
Here, we propose the general subpopulation framework that has the capability of integrating optimization algorithms without restrictions as well as aid the design of structured algorithms. 
The proposed framework is capable of generalizing most of the structured evolutionary algorithms, such as cellular algorithms, island models, spatial predator-prey and restricted mating based algorithms under its formalization.
Moreover, we propose two algorithms based on the general subpopulation framework, demonstrating that with the simple addition of a number of single-objective differential evolution algorithms for each objective the results improve greatly, even when the combined algorithms behave poorly when evaluated alone at the tests. 
Most importantly, the comparison between the subpopulation algorithms and their related panmictic algorithms suggests that the competition between different strategies inside one population can have deleterious consequences for an algorithm and reveal a strong benefit of using the subpopulation framework.\\

The code for SAN, the multi-objective algorithm which has the current best results in the hardest benchmark, is available at the following \href{https://github.com/zweifel/zweifel}{link}.
\end{abstract}



\newtheorem{defi}{{{\bf Definition}}}

\section{Introduction}
\label{sec:intro}

Although particle swarm optimization algorithms, differential evolution and genetic algorithms follow different lines of thought, they can all be seen from the same framework or structure. Not only these types but most of the algorithms in evolutionary computation share the same framework.
They are based on a single population of individuals, which interacts in some form to produce new ones inside the same population. 
To these types of algorithms, it is usually given the name of unstructured EAs or panmictic \cite{sprave1999unified}.

On the other hand, island based models and cellular algorithms achieved relevant improvements, indicating that the evolutionary bioinspiration, when extended to include concepts of subpopulation and neighborhood aspects, can be advantageous \cite{tomassini2005spatially}.
These types of algorithms are called structured models.

Nonetheless, the use of structured algorithms in multi-objective optimization has been under-explored \cite{nebro2006cellular}.
Lately, we researchers start asking ourselves what could be the next step (future research trends) \cite{coello2006evolutionary}, since very simple and effective algorithms were developed, and it is hard to improve them without losing any of their benefits. 
This article tackles this problem from a different perspective. 
Here, we switch the focus from algorithms to frameworks.\footnote{The definition of framework used in this article refers to a basic structure underlying a set of algorithms that is formalized and exemplified, enabling the understanding and analysis of a class of algorithms rather than a single one.}
Moreover, when changing from a panmictic to a structured framework, small and simple changes may give relevant improvements to the algorithms of the state of the art. 

This article proposes the subpopulation framework which has the following features:
\begin{itemize}
\item \emph{Integration Capability} - It allows for the addition of any number of algorithms which are integrated as subpopulations of the framework. Although this feature is not new, for example it was explored similarly in island models \cite{li2008island}, here we show that not only evolutionary algorithms (EAs) but any optimization algorithm can be integrated in this framework. It is not required for these algorithms to be population based either (examples of how this can be constructed are given in Section~\ref{examples_of_gsf}).
\item \emph{General Formulation} -  This framework is a general case for most of the structured approaches including but not limited to cellular algorithms and island based models (Section~\ref{representation}).	
The formalized subpopulation framework also generalizes the panmictic framework, because the panmictic framework is its special case when the number of subpopulations is fixed to $1$ and the ${\cal IM}$ matrix set (that describes the interaction among subpopulations of the proposed framework, further explained in Section~\ref{gsf}) can be ignored. 
\item \emph{State of The Art Solutions} - Experimentally, it was shown that algorithms based on the subpopulation framework can achieve state of the art results (Section~\ref{problems}). 
In fact, results with the Subpopulation Algorithm based on Novelty (SAN) (Section~\ref{NSA}) can be reasonably regarded as one of the most robust algorithms to date in multi-objective optimization, solving different types of problems in bi-objective and many objective settings with excellent results and surpassing the third version of the Generalized Differential Evolution algorithm (in short GDE3, currently one of the most suitable MOEA of the state of the art \cite{durillo2010study}) in most of the tests.
\end{itemize}

Experiments are conducted with two novel algorithms that implement the proposed subpopulation framework. 
These algorithms are developed based on single population ones (panmictic). 
The chosen panmictic algorithms, which were also used for comparison, are the GDE3 and a simple novelty search algorithm called Multi-Objective Novelty Algorithm (which is also a contribution of this article, described in Section~\ref{monsa-sec})
Here, the intention is to choose algorithms as different as possible to show some aspects of the subpopulation framework and its applicability to any type of algorithm. 
Notice that the dissimilarities in the GDE3 and Multi-Objective Novelty Algorithm arise from the fact that the former is objective-based while the later is novelty-based (further explanation of novelty search is given on Section~\ref{novelty_search}). 
In fact, it will be shown that \textit{the differences present in strategies of two or more subpopulations benefits their integration, in contrast with the competition which arises when different strategies are present in a single population}. 

This article shows that simple subpopulations dynamics can give relevant improvements when combined with an algorithm of the state of the art in the proposed framework, demonstrating the strong benefits of the subpopulation framework. Additionally, the competition between different strategies inside the traditional single-population framework can have deleterious consequences for an algorithm.
This is analyzed and verified experimentally in Section~\ref{explanation}.
Such problems confronted by the panmictic algorithms are similar to the ones confronted by the objective-based algorithms when contrasted with novelty-search based algorithms \cite{lehman2010abandoning}, since they are easily trapped in deceptive fitness landscapes.
The solution provided by the subpopulation framework is that the presence of multiple populations with different dynamics will let the algorithm be less sensitive to local optima. 

Finally, this article presents a discussion over an unexpected result, where the experimental results with a combination of three simple subpopulations achieved state of the art quality in the WFG Toolkit \cite{huband2006review} (presented and explained in Sections~\ref{comparison_1} and~\ref{explanation}).

Sections~\ref{subpopulation-algorithms}, \ref{differential-evolution-sec}, \ref{demo} and \ref{novelty_search} review briefly the literature respectively in similar structured EAs, differential evolution in single-objective optimization, differential evolution multi-objective algorithms and novelty search areas. Thereafter, Section~\ref{gsf} proposes the general subpopulation framework.
Section~\ref{gsa} describes two subpopulation algorithms which use as basis the general subpopulation framework. 
Section~\ref{comparison_methodology} presents the methodology used for comparison, Section~\ref{problems} describes the problems' characteristics and shows the results obtained on them.
Lastly, the conclusions are presented in Section~\ref{conclusion}.

\section{Structured EAs}
\label{subpopulation-algorithms}

On one hand, the usual type of EAs pertain to a class of single population algorithms, which we call here single-population framework. But they are also known as panmictic EAs.
On the other hand, there are other algorithms which spread their population into a structure with some defined interrelationship \cite{alba2002parallelism}.
This paper will follow the definition that structured algorithms are any procedure which may have its population formulated with subpopulation groups, with the number of possible non-trivial subpopulation groups necessarily greater than one.
For example, the simple EA can not be seen as a structured algorithm, since the number of possible subpopulation groups can never be formulated as greater than one \cite{goldberg1989genetic}.
Multi-objective ELSA is a local selection algorithm which also cannot be seen as a structured algorithm \cite{menczer2000efficient}.
Note that some procedures, such as the restricted mating, fit in the previous definition of structured algorithms \cite{zitzler1999multiobjective}.
Therefore, restricted mating based algorithms can be seen as structured algorithms (see Section~\ref{representation} for the complete description). 

Parallel EAs are usually examples of structured EAs which are sometimes divided into three classes \cite{gorges1991genetic}, \cite{sprave1999unified}:
\begin{enumerate}
\item Island Model: The basic structure used by this model consists of multiple subpopulations, where a limited amount of genetic information is exchanged between any of them arbitrarily;
\item Stepping Stone Model: In this model a neighborhood relation is defined, where only the adjacent subpopulations can exchange information. Aside from that, it is defined in the same way as the Island Model;
\item Neighborhood Model: A complex single population structure, where individuals interact only with adjacent individuals.
\end{enumerate}
The cellular algorithm \cite{manderick1989} (also called fine grained model or lattice model) for example pertains to the third class.

According to \cite{de2002psfga}, Parallel MOEA models can be divided also into three classes: global parallelization, coarse grain and fine grain. 
Global parallelization does not present any structured population aspect, while coarse grain (also called island GAs) and fine grain (also called cellular GAs) are parallel versions of structured algorithms already mentioned before. 
In \cite{talbi2008parallel}, the classifications of the parallel models differ from the previous three classes, though from a population structure point of view they can still be converted to the previous three classes.

Other types of EAs were also developed where the evolutionary conditions differed from subpopulation to subpopulation. These were called nonstandard structured EAs and they were reviewed by Alba and Tomassini in \cite{alba2002parallelism}.
Another extensive review of single-objective structured EAs can be found in the book of Tomassini \cite{tomassini2005spatially}.

Regarding multi-objective algorithms, there are also some algorithms which are structured. 
To cite some: multi-objective cellular algorithms \cite{nebro2006cellular}, some rudimentary subpopulation algorithms \cite{santos2010node}, \cite{delbem2005main}, spatial predator-prey MOEA \cite{laumanns1998spatial} and multi-colony ant algorithms \cite{iredi2001bi}. 
Spatial predator-prey MOEA defines an adjacency matrix with edges as solutions where the predator makes a random walk and erases the worst solution in the neighborhood which is related to a given objective \cite{laumanns1998spatial}.
The number of predators walking are as much as there are objectives.
Ant colony optimization algorithms construct a population of solutions by sampling from a probabilistic model (usually in the form of a matrix of pheromone).
This matrix of pheromone is constantly updated by the ants.
Although they can not be defined as structured algorithms by the definition above, their multi-colony version can be defined.
Multi-colony optimization algorithms use normally multiple matrices of pheromone with some rules to decide how and which pheromone matrix to be updated/used.

Moreover, a generalized framework of the structured algorithms is still non existent. 
This article fills this gap by formalizing a general unifying framework capable of representing most if not all of these structured models.

\subsection{Related Methods}
\label{rel_methods}

Although being a single population algorithm, AMALGAM is related to the proposed framework since both can be used to integrate algorithms. 
AMALGAM is a panmictic multiobjective algorithm that create a number of offspring points using genetic operators from different algorithms.
Fast nondominated sorting is used to rank the offsprings together with the previous population, subsequently defining the next population \cite{vrugt2007improved}.

As told before, one important difference between AMALGAM and the proposed framework is that the first is panmictic.
Therefore, it has the disadvantage that multiple algorithms joined together may conflict with each other in the single pool of solutions.
Another important difference is that AMALGAM can only define the integration of algorithms with biological models for population evolution, since genetic operators are necessary for the integration.
Here, the proposed framework define the integration of any optimization algorithm.


The portfolio design proposed by \cite{gomes1997algorithm} runs different algorithms (strategies) or copies of the same strategy with the objective of selecting the best strategy for the given problem.
Details of how the selection and evaluation of strategies as well as the strategies themselves are dependent on the problem at hand \cite{guerri2004learning}.
The strategies run without communication between each other. 
Therefore, when considered under the light of the framework described here, the set of interaction matrices is null and can be ignored (interaction matrices are part of the frameword defined in Section~\ref{gsf}).
The similarities between this method and the proposed framework are limited to the use of multiple algorithms together.

\section{Differential Evolution}
\label{differential-evolution-sec}

The Differential Evolution (DE) is a meta-heuristic contained in the subfield of evolutionary computation, which can be employed for optimizing multi-dimensional real-value functions, where these functions are neither required to be continuous nor to be differentiable. 
It solves problems using a simple algorithm similar to the ones used by EAs, but the operators used by the DE are not based on the evolution of species \cite{storn1997differential}.  
The algorithm is described succinctly in Table~\ref{de_alg} and the procedures of mutation, crossover and selection are explained in the following subsections.
\begin{table}[h]
 \centering
\caption{Differential Evolution Algorithm} 
\begin{tabular}{p{12cm}}
\hline
\begin{enumerate}
\item Initialize population with random samples uniformly distributed over the search space
\item Repeat for each individual until a criterion of convergence is met
\begin{enumerate}
\item Mutation
\item Crossover
\item Selection
\end{enumerate}
\item Return solution
\end{enumerate} \\
\hline
\end{tabular}
\label{de_alg}
\end{table}

\subsection{Mutation}

For each vector $x_{i,g}$, where $i$ is the index of this vector (which relates to the individual index in the population, since each individual has its own vector) and $g$ is the current generation where the vector takes place, the mutation is applied by creating a mutant vector based on a numerical operator described in Equation~\ref{de_mutation_eq}.
\begin{equation}
v_{i,g+1} = x_{r1,g} + F(x_{r2,g}-x_{r3,g}),
\label{de_mutation_eq}
\end{equation}
where $r1$, $r2$ and $r3$ are randomly selected individuals of the population, which must differ from the individual $i$. $F$ is a parameter which should meet the condition $F \in [0,2]$. 

\subsection{Crossover}

During the crossover, trial vectors $u_{i,g+1}$ are created from a combination of the mutation vector $v_{i,g+1}$ and the original vector $x_{i,g}$. 
The trial vector created is expressed in Equation~\ref{de_crossover_eq}.
\begin{equation}
u_{i,j,g+1} = \left\{\begin{array}{cc}
 x_{i,j,g} & \mbox{if $rand() > CR$ and $j \neq rnd_i$};\\
 v_{i,j,g+1} & \mbox{if $rand() \leq CR$ or $j = rnd_i$},
\end{array} \right. 
\label{de_crossover_eq}
\end{equation}
where $rand() \in [0,1]$ is an uniformly distributed random number, $CR \in [0,1]$ is a parameter passed to DE, $j$ is the vector component index and $rnd_i$ is a randomly chosen index, with the objective of choosing at least one component from the vector $v_{i,j,g+1}$.

\subsection{Selection}

The selection is the last step of the generation, where it is determined for each vector if the trial vector $u_{i,g+1}$ will substitute the original vector $x_{i,g}$ or not. 
For this, both the $u_{i,g+1}$ and $v_{i,g}$ are evaluated and the vector with better fitness function is kept, forming the next generation vector $x_{i,g+1}$.

\subsection{Comparison with other Evolutionary Algorithms}

The DE algorithm and its variations are known by their robustness, quality of the solutions, short running time, easy use and application to a wide range of applications not limited by the type of the objective function \cite{storn1997differential}, \cite{brest2006self}. 

Promising results were obtained in numerous different experiments. 
Two variations of it achieved the best solutions on all problems from ICEC'96 \cite{storn2002minimizing}. 
In the work of \cite{vesterstrom2004comparative} it was shown to achieve more accurate solutions, faster and with greater robustness than Particle Swarm Optimization (PSO) and Evolutionary Algorithms (EAs). At the state of the art, the DE is still compared on equal grounds to complex optimization algorithms (e.g., Estimation of Distribution Algorithms) \cite{garcia2009study}.

\section{Differential Evolution based Multi-objective Methods}
\label{demo}

The DE was shown to achieve significant improvements over other single-objective \cite{vesterstrom2004comparative} as well as in multi-criteria optimization algorithms \cite{tuar2007differential}, \cite{durillo2010study}. 
The reason behind this overall better results lies partially on the rotational invariant behavior of DE's operators, which adapts to the fitness landscape when compared with NSGA-II's genetic operators and other algorithms with similar genetic operators \cite{iorio2005solving}.
Recent studies show that in multi-objective-problems, DE is one of the best approaches when the problem size increases in scale \cite{durillo2010study}.

There are various multi-objective methods based on differential evolution \cite{chakraborty2008advances}. 
They can be divided into old versions of algorithms which used only Pareto dominance to select individuals and modern methods which use the Pareto dominance and a diversity measure for selection \cite{tuar2007differential}.
It is generally accepted that the last version of the generalized evolution algorithm (the GDE3 \cite{kukkonen2005gde3}) and the differential evolution multi-objective algorithm (DEMO) \cite{robi2005demo} are the representatives of the modern class of multi-objective algorithms based on DE \cite{tuar2007differential}, \cite{durillo2010study}.
By taking into account that DEMO \cite{robi2005demo} is also similar to GDE3 \cite{kukkonen2005gde3} algorithm, without constraint handling and a fallback to the original DE in the case of single-objective, we will conduct the comparison and study on GDE3 solely.

Recently, a comparison between eight modern multi-objective algorithms was made \cite{durillo2010study}. 
They showed evidences that GDE3 is currently one of the most suitable MOEA of the state of the art.
Among the results, it is stated that GDE3 tends not only to be faster, but also scales better in relation to the number of decision variables. In the tests made, there was only one other algorithm of the state of the art based on the PSO approach with similar performance.

\subsection{General Differential Evolution 3}
\label{gde3}

GDE3 has the same basic loop as the DE, with a modification in the selection phase and an addition of a pruning stage.
In the selection phase, the algorithm considers the Pareto dominance and the constraints. Let $s$ and $t$ correspond respectively to the solution and its respective trial solution. Then, in the selection phase the following statements apply:
\begin{itemize}
\item If both $s$ and $t$ are not feasible. The trial solution $t$ substitute $s$ only when it dominates the solution $s$ in unconstrained space;
\item If one solution is feasible and the other is not feasible, the feasible solution is chosen;
\item Finally, if both solutions are feasible, the solution which dominates the other is kept. However, if neither one dominates the other, both solutions are added to the next population, increasing the size of the population.
\end{itemize}

As a consequence of the modifications in the selection stage, the pruning stage was added to keep the population to a minimum because GDE3's selection phase described above can make the population increase in size. 
The pruning stage consists of sorting based on a diversity measure, consecutively selecting the first individuals to fill the next population size. 

In the first version, GDE3 used crowding distance as its diversity measure \cite{kukkonen2005gde3}, similar to NSGAII \cite{deb2002fast}. But in its most recent version the $k$-nearest neighbors measure was used as a distance measure. 
This metric was shown to be more consistent than the crowding distance measure when the number of objectives is greater than two \cite{kukkonen2007performance}.
The experiments conducted in this paper use GDE3 with a $k$-nearest neighbors measure.

\section{Novelty Search}
\label{novelty_search}

In nature, evolution is usually observed as an open-ended process which continually creates individuals with greater complexity and diversity \cite{maley1999four}. 
Novelty search is a method developed by Lehman and Stanley that mimics the open-ended evolutionary process with a simple novelty metric \cite{lehman2010abandoning}, \cite{lehman2008exploiting}, rewarding novel individuals with a direct measure of novelty.

Moreover, in the perspective of optimization, problems are sometimes deceptive. This is usually the case for real world problems, because when problems increase in size and complexity it is improbable that a fitness function exists which can drive the algorithm directly to the goal. 
Novelty search aids the optimization in these deceptive spaces by identifying stepping stones, which are the novel individuals found by the novelty metric.

Recently, novelty search was used in very distinct areas such as neuro-evolution \cite{mouret2009using}, \cite{lehman2010abandoning}, genetic programming \cite{lehman2010efficiently}, multi-objective evolution \cite{mouret2009using} and robotics \cite{doncieux2010behavioral}, \cite{doncieux2009exploring}.
Moreover, there are an ever increasing number of articles with further evidence of novelty search benefits in deceptive problems.
Some papers even showed the astonishing find that novelty search can be used sometimes as a substitute of objective-based search \cite{lehman2010abandoning}, \cite{woolley2011deleterious}. The good results of novelty search in relation to objective-based search revealed that objective-based search may have deleterious effects on search.

\subsection{Novelty Metric}
\label{novelty_metric}

For measuring the novelty of a solution, the novelty search relies on a metric which can be any equation capable of describing how much an individual is novel in comparison with the past individuals of the archive.
The usual metric used is the $k$-nearest neighbors which was also employed by Lehman and Stanley in their pioneering work on novelty search\cite{lehman2008exploiting}. 
The following equation defines it exactly:
\begin{equation} 
p(x)=\frac{1}{k}\sum^{k}_{i=1}dist(x,\mu_i),
\end{equation}
where $k$ is a parameter defined arbitrarily, $\mu_i$ is the $i$-th nearest neighbor of $x$ according to the distance measure $dist()$. 
The distance measure is problem dependent. 
Usually, it is calculated in the behaviors space rather than fitness space, where behaviors space is composed as the small set of features which identifies a unique behavior (reducing the search space and differing in this way from exhaustive enumeration).
The archive is an incremental set of individuals, receiving new individuals only if they surpass a novelty threshold $n_{min}$ adjusted automatically by some rule.

It goes often unnoticed, but one of the problems of this novelty metric lies on its dynamic adjustment, i.e., the parameters used to update the archive.
The following are the dynamics commonly used to update the metric:
\begin{itemize}
\item if more than $n_a$ individuals entered the archive, multiply $n_{min}$ by $n_{inc}$;
\item if $n_r$ individuals did not enter in the archive, multiply $n_{min}$ by $n_{dec}$;
\end{itemize}
where $n_a$, $n_r$ are positive integers (refers to the number of individuals), 
$n_{inc}$, $n_{dec} \in \mathbb{R}, n_{inc} > 1, 0 <  n_{dec} < 1$ (refers to values of the novelty metric).
These parameters define the rate of individuals which enter the archive. 
It follows that the bigger the archive is the more sensitive the novelty metric is to identify new individuals, because the higher the number of points, the less separated the points will be from each other. Then, a bigger archive is a direct result from a smaller $n_{min}$ and consequently a more sensitive search with less chances of letting new individuals go unnoticed.
On the other hand, a bigger archive makes the metric evaluation slower.

\subsection{Multi-objective Novelty Algorithm (MONA)}
\label{monsa-sec}

In this Section, we propose MONA.
The first algorithm to use novelty in a multi-objective context. 
The algorithm uses solely novelty search.
Therefore, this algorithm follows the same line as the Lehman and Stanley study \cite{lehman2010abandoning}, hypothesizing that an algorithm based on the novelty alone might be better than objective based methods.
MONA is a very simple algorithm proposed in this article, where the space of all the objectives is taken to be the behavior space of the novelty, differently from the Mouret approach \cite{mouret2009using} where novelty was seen as an additional objective. 
Table~\ref{monsa} describes the algorithm.

\begin{table}[h]
 \centering
\caption{Multi-Objective Novelty Algorithm} 
\begin{tabular}{p{12cm}}
\hline
\begin{enumerate}
\item Initialize population with random samples uniformly distributed over the search space
\item Repeat for each individual in the population until a criterion of convergence is met
\begin{enumerate}
\item Apply the same mutation and crossover operators as used by DE
\item Calculate the novelty metric
\item Verify if its novelty metric is above the $n_{min}$ threshold, if it is above insert it on the archive (unlimited in size)
\item Update $n_{min}$ (see Section~\ref{novelty_metric})
\item Create a new population by sampling uniformly with replacement from the archive
\end{enumerate}
\item Return the archive's non-dominated solutions as the solution set
\end{enumerate} \\
\hline
\end{tabular}
\label{monsa}
\end{table}

The purpose of this algorithm is to be a very simple algorithm, which will be compared as well as used in the general subpopulation framework, showing that from very simple bases efficient and robust algorithms can be constructed.

\section{General Subpopulation Framework}
\label{gsf}

The General Subpopulation Framework (GSF) is proposed here as an underlining structure of a class of multi-objective algorithms which unifies a number of structured EAs in its formalization.
Additionally, it is capable of integrating different optimization algorithms without restrictions. 
This flexible ability of joining algorithms together is important as it will be shown in the experiments. 
Mostly, because this type of cooperation between algorithms can sum their benefits while the competition between them in each subpopulation is decreased to a minimum.

In this context, we define:
\begin{defi}
\label{subpop_definition}
\emph{Subpopulation\\ 
A subpopulation is a finite set of individuals related with a group of well defined dynamics. 
These dynamics are usually (although not necessarily) composed of interactions of these individuals with either themselves or individuals of other subpopulations.
But they are not in any way limited to it. 
}
\end{defi}

When connecting these subpopulations together, a new matrix appears. 
To this matrix is given the name ${\cal IM}$. It is formally defined as follows:
\begin{defi}
\label{im_definition}
\emph{${\cal IM}$ - Subpopulation Interaction Probability Matrix Set\\ 
The subpopulation interaction probability matrix set ${\cal IM}$ is a set of matrixes of the form:}
\begin{equation}
{{\cal IM}} = \{IM_1, IM_2, ..., IM_m\},
\end{equation}
\emph{where $m$ is the number of types of interactions used in an optimization algorithm. And each $IM_i$ corresponds to the following matrix: }
\begin{equation}
IM_i =
\begin{pmatrix}
  p_{i,1,1} & p_{i,1,2} & \cdots & p_{i,1,s} \\
  p_{i,2,1} & p_{i,2,2} & \cdots & p_{i,2,s} \\
  \vdots  & \vdots  & \ddots & \vdots  \\
  p_{i,s,1} & p_{i,s,2} & \cdots & p_{i,s,s}
\end{pmatrix},
\end{equation}
\emph{where $s$ is the number of subpopulations and $p_{i,a,b}$ is the probability of an interaction $i$ occurring in subpopulation $a$ and taking as parameters the individuals of subpopulation $b$ or the subpopulation $b$ itself.
}
\end{defi}

The evolutionary operators are examples of interactions. For example in the case of a subpopulation based version of DE's operators, let us assume their interactions are described by $IM_d$. Then, the trial vector would be, for each individual of this subpopulation, composed of three individuals chosen based on the probabilities of the $IM_d$ matrix.
Recall that the ${\cal IM}$ matrix set can be ignored in the case of only one subpopulation and this is why it can be ignored for  panmictic algorithms.

Notice also that the interaction of each subpopulation can also differ from subpopulation to subpopulation. In the case of just one subpopulation $k$ having an interaction~$i$, $IM_i$ would be of the following form:
\begin{equation}
IM_i =
\begin{pmatrix}
  0 & 0 & \cdots & 0, \\
  \vdots  & \vdots  & \ddots & \vdots  \\
  p_{i,k,1} & p_{i,k,2} & \cdots & p_{i,k,s} \\
  \vdots  & \vdots  & \ddots & \vdots  \\
  0 & 0 & \cdots & 0
\end{pmatrix}.
\end{equation}
Naturally, more complicated global dynamics might also be present, such as dynamical probabilities that depend on time $t$:
\begin{equation}
IM_i =
\begin{pmatrix}
  p_{t,i,1,1} & p_{t,i,1,2} & \cdots & p_{t,i,1,s} \\
  p_{t,i,2,1} & p_{t,i,2,2} & \cdots & p_{t,i,2,s} \\
  \vdots  & \vdots  & \ddots & \vdots  \\
  p_{t,i,s,1} & p_{t,i,s,2} & \cdots & p_{t,i,s,s}
\end{pmatrix}.
\end{equation}


Additionally, the population size variable is extended to a vector version. 
Because the proposed framework has a number of subpopulations, each with a given size.
This vector is hereby called ${\cal S}$ and is defined as follows:
\begin{defi}
\label{definition_bbi}
\emph{${\cal S}$ - Vector of Subpopulation Sizes \\ 
The subpopulations' sizes are defined by vector ${\cal S}$, which corresponds to:}
\begin{equation}
S = ( \check{np_{1}}, \check{np_{2}}, ..., \check{np_{s}} ), \\
\end{equation}
\emph{where $\check{np_{a}}$ is the size of subpopulation $a$. The \textit{total subpopulation size} ($ts$) is naturally:
\begin{equation}
ts = \sum_{j=1}^s\check{np_{j}}. \\
\end{equation}
An equivalent and more convenient representation exists which is independent of the total subpopulation size. 
Let $np_a=\frac{\check{np_a}}{ts}$, corresponding to the ratio of the total subpopulation.
Then, the following representation is also verified:
\begin{equation}
\begin{split}
&np_{a}\in\{x \in \mathbb{R} : 0 < x < 1\} \\
&\sum_{j=1}^{s}np_{j}=1.
\end{split}
\end{equation}
}
\end{defi}

With the previous definitions it is possible to describe explicitly the GSF:
\begin{defi}
\label{definition_bbi2}
\emph{GSF - General Subpopulation Framework \\ 
Suppose we have $s$ subpopulations, then ${\cal P}$ is the set of subpopulations ${\cal P}=\{P_1, ..., P_s\}$ and ${\cal A}$ is the set of panmictic algorithms ${\cal A}=\{A_1, ..., A_s\}$ where a subpopulation $P_i$ is constructed by an algorithm (strategy) $A_i$. 
Therefore, the GSF is defined as a 4-tuple $<{\cal P}, {\cal A}, {\cal S}, {\cal IM}>$, where ${\cal S}$ and ${\cal IM}$ were previously defined as respectively the vector of subpopulation sizes and the set of interaction probability matrices.
}
\end{defi}

The subpopulations may even be used to join arbitrary algorithms which may not even be based on populations.
That is, as long as each algorithm can generate a set of solutions to compose the subpopulation which is representative of its dynamics, the subpopulation framework can handle the joining process (examples are given in Sections~\ref{examples_of_gsf} and~\ref{gsa}). 
For example, in the case of the random search algorithm the subpopulation can be constructed from the last generated solutions.
Therefore, to the knowledge of the authors, any algorithms can be joined (mixed) by using this framework.
Naturally, for the inclusion of an algorithm in this framework it is also relevant but not necessary to have:
\begin{itemize}
 \item Dynamics taking into account different individuals of its population (which can be modified to handle any individual of any population by the ${\cal IM}$ set of matrices);
 \item Different dynamics from the other subpopulations present in the framework. This can be relevant, since the higher the similarities between subpopulations are the less important the subpopulations become, in other words, multiple subpopulations with similar dynamics will produce results similar to a single population.  
\end{itemize}

The following subsections demonstrate how GSF can represent most of the optimization algorithms. 
Section~\ref{representation} shows how GSF can represent various types of structured EAs, while Section~\ref{examples_of_gsf} gives two examples of famous algorithms (one a panmictic EA and the other a non-evolutionary algorithm) as well as shows how they can be transformed to the GSF approach without losing many of their characteristics.
In Section~\ref{gsa}, two new algorithms are proposed based on their related panmictic algorithms. This time, however, the objective is not merely illustrative, since the algorithms described possess important features described in detail later on.
In fact, these important features enable them to surpass algorithms of the state of the art. 

\subsection{Representation Capabilities}
\label{representation}

The general subpopulation framework can represent various types of structured EAs, including:
\begin{itemize}
\item Island-Based Models\cite{tomassini2005spatially} - Each panmictic island forms a subpopulation $P_i$ with the set of algorithms ${\cal A}$ containing identical algorithms for all subpopulations. 
Let the number of panmictic islands be $s$, then ${\cal S}=(\frac{1}{s},...,\frac{1}{s})$ and $|A|=|P|=s$.
Between the subpopulations an interaction defined by the exchange of genetic information can be formalized with an $IM_1$ matrix of the form:
\begin{equation}
IM_1 =
\begin{pmatrix}
  0 & p_{i,1,2} & \cdots & p_{i,1,s} \\
  p_{i,2,1} & 0 & \cdots & p_{i,2,s} \\
  \vdots  & \vdots  & \ddots & \vdots  \\
  p_{i,s,1} & p_{i,s,2} & \cdots & 0.
\end{pmatrix}
\end{equation}
That is, each individual selected for exchange must necessarily go to another subpopulation, therefore the diagonal entries are zero.
This dynamic is usually the unique one which other subpopulations can participate in. Inside the algorithms other dynamics can take place (e.g., crossover) and these would have also a trivial set of $IM_i$s with the only non-null probabilities residing on its diagonal (i.e., the interactions happen only inside the same subpopulation).
\item Cellular Algorithms\cite{manderick1989} - This type of algorithm can be thought as the opposite line of thought in comparison with Island-Based Models, where the number of subpopulations is maximized with the minimum possible size of subpopulations, i.e., cellular algorithms can be seen as a large number of subpopulations $P_i$ of equal size $1$.
Let the number of cells in a given cellular algorithm be $s$, then ${\cal S}=(\frac{1}{s},...,\frac{1}{s})$ and $|P|=s$ with each individual cell corresponding to a subpopulation $P_i$ and the update of each cell can be divided into $s$ algorithms forming the ${\cal A}$ set of panmictic algorithms.
Consider the case of a cellular algorithm with nine individuals with a von Neumann neighborhood, then it possesses nine subpopulations and an $IM_c$ matrix defined by:
\begin{equation}
IM_c =
\begin{pmatrix}
  0 & \frac{1}{4} & \frac{1}{4} & \frac{1}{4} & 0 & 0 & \frac{1}{4} & 0 & 0 \\
  \frac{1}{4} & 0 & \frac{1}{4} & 0 & \frac{1}{4} & 0 & 0 & \frac{1}{4} & 0 \\
  \vdots  & \vdots  & \vdots & \vdots  & \vdots  & \vdots & \vdots  & \vdots  & \vdots \\
  0 & 0 & \frac{1}{4} & 0 & 0 & \frac{1}{4} & \frac{1}{4} & \frac{1}{4} & 0  
\end{pmatrix}.
\end{equation}
Moreover, all interactions of cellular algorithms use the same neighborhood, therefore the set of matrices ${\cal IM}$ is given by:
\begin{equation}
{\cal IM} = \{IM_c, IM_c, ..., IM_c\}.
\end{equation}
In some certain cellular algorithms, a dynamical $IM_c$ has to be used to represent the change of neighborhood of each cell.
\end{itemize}

\begin{itemize}
\item Restricted Mating \cite{zitzler1999multiobjective} - 
Some procedures although not related to subpopulations at first glance, can be converted to this formalization.
Restricted mating, for example, can be formalized with subpopulations. 
By considering each subpopulation containing only one individual, we have the restricted mating interaction defined by:
\begin{equation}
IM_1 =
\begin{pmatrix}
  0 & p_{1,2} & \cdots & p_{1,s} \\
  p_{2,1} & 0 & \cdots & p_{2,s} \\
  \vdots  & \vdots  & \ddots & \vdots  \\
  p_{s,1} & p_{s,2} & \cdots & 0
\end{pmatrix},
\end{equation}
when for any $(a,b)$ pair, $p_{a,b}$ becomes:
\begin{equation}
 u_{a,b} = \left\{\begin{array}{cc}
 1 & \mbox{if $dist(a,b) < \sigma$};\\
 0 & \mbox{otherwise},
\end{array} \right. \\
\end{equation}
\begin{equation}
p_{a,b}= \frac{u_{a,b}}{\sum_{i=1}^{s}u_{a,i}}
\end{equation}
$\sigma$ is an arbitrary threshold and $dist(a,b)$ is usually the Euclidean distance between solutions $a$ and $b$ \cite{zitzler1999multiobjective}.

\item Spatial Predator-Prey MOEA \cite{laumanns1998spatial} - This algorithm defines an adjacency matrix $G$ with edges as solutions where the predator makes a random walk.
This algorithm can be reformulated into the subpopulation framework by considering as interaction the replacement of the preys selected by the predators. 
Although the replacement can be done of multiple ways, only the edges in the predator's neighborhood participate.
Therefore, for the replacement interaction, each position $(x,y)$ of the interaction matrix becomes:
\begin{equation}
IM_1(x,y) = min\{G(k,x),G(k,y)\},
\end{equation}
where $k$ is the edge of the predator responsible for this interaction matrix.
Basically, two solutions can only interact if they are in the $k$ (predator's edge) neighborhood.

\item Multi-colony Ant Algorithms - Ant colony optimization algorithms in general are difficult to map into the subpopulation framework because they use population models instead of the solutions themselves. 
This problem is faced similarly when trying to convert estimation of distribution algorithms \cite{pelikan2005bayesian}, \cite{larranaga2002estimation}.
Additionally, some of these methods do not possess a population structure. For example, ant colony optimization algorithms with one colony do not use a structure approach to optimization following the definition above, i.e., although the construction of the solutions by the ants use solution components organized in a structured way, the population of solutions itself is not structurally formulated \cite{iredi2001bi}.
However, some of them such as the multi-colony ant algorithms do have a population structure. 
In this case, it is possible to approximate roughly the population model (e.g., the pheronomone matrix) as a subpopulation and consider the interrelation between them as interactions with their respective interaction matrices.
That is, the pheromone matrices update interaction can be represented as:
\begin{equation}
IM_1 =
\begin{pmatrix}
  1 & 0 & \hdots & 0 \\[0.3em]
  0 & 1 & \hdots & 0 \\[0.3em]
  \vdots & \vdots & \ddots & \vdots \\[0.3em]
  0 & 0 & \hdots & 1
\end{pmatrix},
\end{equation}
when the update is only realized at the original colony.
And when the update is done by region ($\{L_1, L_2, ..., L_s\}$) in the nondominated front, for a given solution $a$ we have: 
\begin{equation}
IM_1 =
\begin{pmatrix}
  a \in L_1 & a \in L_2 & \hdots & a \in L_s \\[0.3em]
  a \in L_1 & a \in L_2 & \hdots & a \in L_s \\[0.3em]
  \vdots & \vdots & \ddots & \vdots \\[0.3em]
  a \in L_1 & a \in L_2 & \hdots & a \in L_s \\[0.3em]
\end{pmatrix}.
\end{equation}

\end{itemize}

\subsection{Examples of Panmictic to GSF Conversion}
\label{examples_of_gsf}

This subsection shows how optimization algorithms of almost any type can be converted to multi-population versions represented by the GSF. 
Examples of both the simple genetic algorithm \cite{goldberg1989genetic} and the simulated annealing \cite{kirkpatrick1983optimization} will be presented. Their ${\cal IM}$ matrix sets will be defined and, among other things, it will be shown how their dynamics could be used to affect other subpopulations.
Notice that the conversions will not make explicit the vector of subpopulations sizes ${\cal S}$, since this parameter is not related with the representation and thus it can be established independently. 

\subsubsection{Simple Genetic Algorithm}
There are three basic procedures in a simple genetic algorithm: crossover, mutation and selection.
However, mutation does not depend on other individuals and selection is executed over a set of individuals of its own population.
Then, it does not make sense to define an interaction matrix for them. 
The mutation and selection can be normally applied, with the only difference from the single population version being that the target is now the current subpopulation (i.e. not the entire population). 
In fact, this slight modification defines the algorithm $A_i$ which constructs its respective subpopulation $P_i$ under the GSF formulation.

Thus, let us define the ${\cal IM}$ set, which consists of only the crossover interaction (${\cal IM}=\{IM_1\}$). 
The crossover interaction matrix $IM_1$ defines the probabilities that an individual of a given subpopulation participate in the crossover.
The exact values of the $IM_1$ is the trivial $IM_1={1}$. 
Note that the simple GA is not a structured algorithm (there are not any other subpopulation to interact with).
However, the designer might want to modify $IM_1$ when joining this algorithm with other algorithms.


\subsubsection{Simulated Annealing}

One of the main difficulties that can be spotted on the simulated annealing is that it is not a population-based algorithm.
This problem can be circumvented by adding the recent modifications of the variables' values in a First In First Out data structure, creating a subpopulation derived from its dynamics. 
Therefore, the simulated annealing algorithm plus the creation of a subpopulation defines algorithm $A_i$ to be applied on its created subpopulation $P_i$.

Lastly, the interaction matrices are defined by an empty set (${\cal IM}=\{\}$), since there is no interaction between solutions in its dynamics.
An empty ${\cal IM}$ might be unappealing at first glance, but when joined with the subpopulations of other algorithms, the subpopulation constructed by this algorithm might be used by other interactions and consequently influence the global dynamics.

\subsection{What is the benefit of using GSF to describe a panmictic algorithm?}

It was shown before that panmictic algorithms can be converted to the GSF. 
However, they possess a trivial $IM$ and bring little explanation. 
Thus, one might question about the usefulness of such a conversion.

The answer is that, once converted to the GSF, any panmictic algorithms can be integrated seamlessly as a subpopulation in other GSF based algorithms.
Section~\ref{gsa} will show some examples of algorithms constructed using the GSF.

Last but not least, the pressures of different panmictic algorithms can be compared by weighting their subpopulations' sizes.
Comparison of algorithms is an important and complicated subject which is aided by GSF. 
GSF also enables a relatively easy evaluation of the cooperation between algorithms, facilitating the construction of hybrid algorithms with the simple addition or deletion of subpopulations.

\subsection{What is the benefit of using GSF?}

One feature of the subpopulation framework is the division of interactions over interaction matrices.
Thus, one can separate only the interactions under interest and compare their structural behavior by looking at those matrices.
For example, it is possible to see that both spatial predator-prey and cellular algorithms are similar in the sense that both use similar interaction matrices (neighborhood matrices).

Moreover, designing structured algorithms may become easier by looking at different interactions and interaction matrices instead of multiple structures and their internal behavior.
The framework also aids other abstractions such as a mix between structures (i.e., sometimes the structure behave like a cellular algorithm and sometimes like a island model) by the simple inclusion of other interaction matrices.
For example, the inclusion of a cellular's interaction matrix into a island model algorithm.


\section{Evaluation of General Subpopulation Algorithms (GSAs)}
\label{gsa}

To evaluate the subpopulation framework appropriately, we elaborate two subpopulation algorithms: one based on GDE3 (see Section~\ref{gde3}) and the other based on MONA (see Section~\ref{monsa-sec}).
We will hereby call these GSAs respectively the Subpopulation Algorithm based on General Differential Evolution (SAGDE) and the Subpopulation Algorithm based on Novelty (SAN).

\begin{figure}[h!]
 \centering
 \includegraphics[scale=0.45]{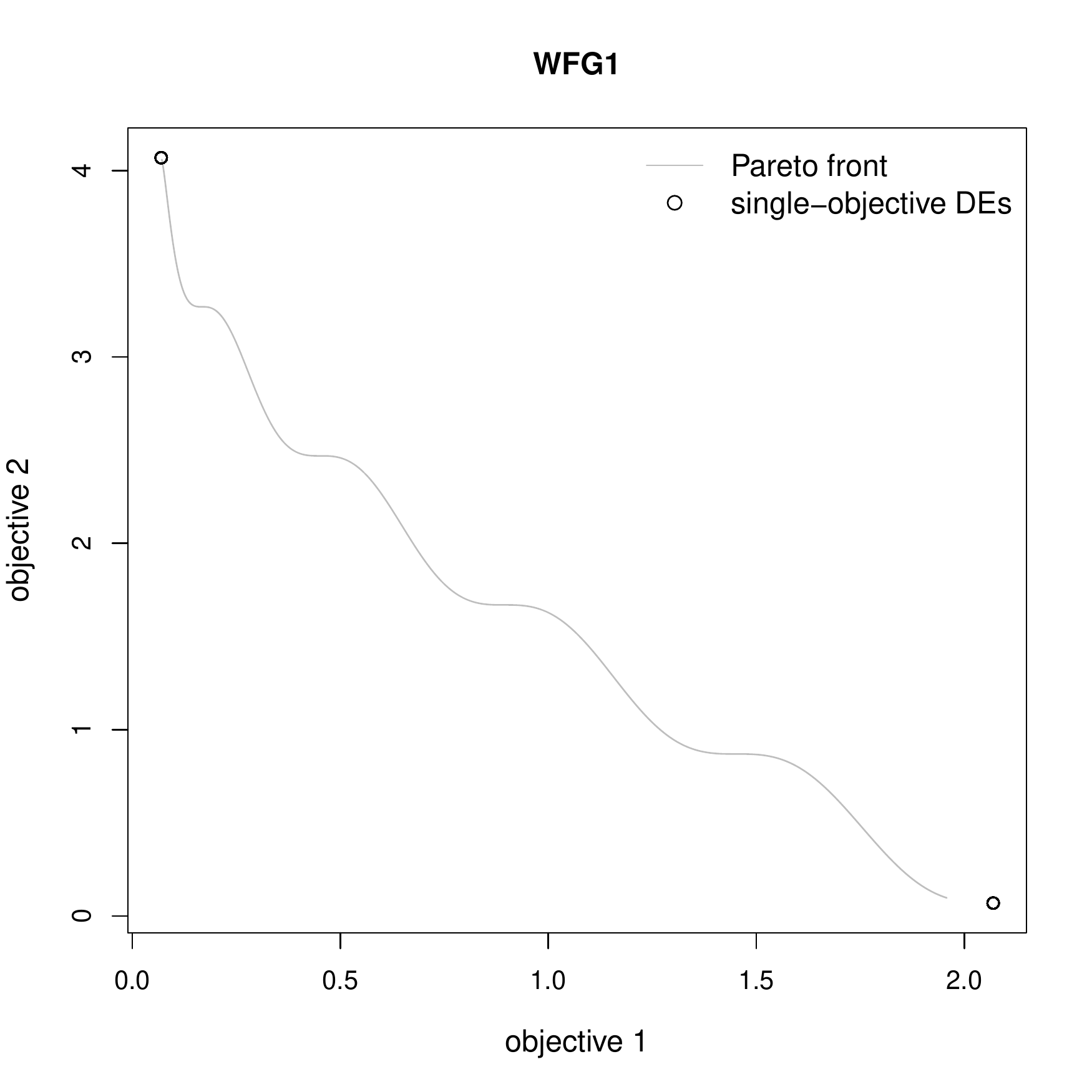}
 \caption{Solutions of single-objective DEs, each one evolving a different objective. For the test, the WFG1 problem is used with $2$ objectives, $20$ distance parameters and $4$ position parameters. Each DE had a population of $50$ individuals, with $CR=0.6$, $F=0.5$, and maximum number of generations of $25000$.}
 \label{des_only}
\end{figure}

Both SAGDE and SAN are motivated by the fact that single-objective DEs evolved at each objective usually achieve good results.
Take for example the WFG1 problem \cite{huband2006review}.
If we apply a GSA made uniquely of subpopulations of single-objective DEs, each evolving a different single objective, we achieve usually the result plotted in Figure~\ref{des_only}. 
Note that the DEs achieve good results on each single objective, with the resultant individuals very close to the Pareto front, but the front is hardly covered. 
Then, what if another subpopulation is added to this algorithm, which might wisely ``mix" these DEs solutions?
The following algorithms are motivated by this question and in Section~\ref{problems} an extensive answer is given based on the experiments.

\subsection{SAGDE}
\label{GDESA}

In a problem with $n$ objectives, SAGDE has $n+1$ subpopulations ${\cal P} = \{P_1,...,P_{n+1}\}$, where $\{A_1,...,A_n\}$ are single-objective DEs with each one evolving a different objective and the GDE3 (multi-objective algorithm) is used as the algorithm $A_{n+1}$ for the subpopulation $P_{n+1}$.
The GDE3 subpopulation as well as the $n$ single-objective differential evolution subpopulations behave in the same way as usual aside from the fact that an uniform matrix $IM_1$ (shown in Equation~\ref{uniform_im}) is used to determine which individual will be part of the trial vector in the differential operator, i.e., ${\cal IM}=\{IM_1\}$ where:
\begin{equation}
IM_1 =
\begin{pmatrix}
  \frac{1}{n+1} &\frac{1}{n+1} & \hdots & \frac{1}{n+1} \\[0.3em]
  \frac{1}{n+1} &\frac{1}{n+1} & \hdots & \frac{1}{n+1} \\[0.3em]
  \vdots & \vdots & \ddots & \vdots \\[0.3em]
  \frac{1}{n+1} &\frac{1}{n+1} & \hdots & \frac{1}{n+1} 
\end{pmatrix}.
\label{uniform_im}
\end{equation}

\subsection{SAN}
\label{NSA}

In the same way as SAGDE, SAN has $n+1$ subpopulations ${\cal P} = \{P_1,...,P_{n+1}\}$ with $n$ of them made of single-objective DEs ($\{A_1,...,A_n\}$), where $n$ is the number of objectives of the problem. 
Each single-objective DE optimizes a different objective and there is an additional subpopulation corresponding to the MONA ($A_{n+1}$) (multi-objective algorithm based on the novelty search approach proposed by this article, see Section~\ref{monsa-sec}).

Both MONA and the $n$ single-objective DE subpopulations behave in the same way as usual with the unique differences being the use the same uniform matrix $IM_1$ described in Equation~\ref{uniform_im} (i.e., an individual chosen has an uniform probability of $\frac{1}{n+1}$ of coming from any subpopulation) to select individuals for the trial vector in the DE operator used in both algorithms. 
Moreover, MONA verifies any new individuals generated by any subpopulation for inclusion in the novelty archive (i.e., not only its own generated individuals).
In other words, the inclusion of solutions in the novelty archive is a different interaction defined by $IM_2$.
It is activated every time a new solution is created in any subpopulation, $IM_2$ matrix is defined below:
\begin{equation}
IM_2 =
\begin{pmatrix}
  0 & 0 & \hdots & 1 \\[0.3em]
  0 & 0 & \hdots & 1 \\[0.3em]
  0 & \vdots & \ddots & \vdots \\[0.3em]
  0 & 0 & \hdots & 1 
\end{pmatrix},
\label{im_matrix2}
\end{equation}
where the last column is referent to the MONA's subpopulation.

\section{Comparison Methodology}
\label{comparison_methodology}

To compare algorithms the following procedure is used:
\begin{enumerate}
\item Realize multiple runs of the algorithm and store the solution sets.
\item For each solution set do:
	\begin{itemize}
		\item Compute the hypervolume indicator (Section~\ref{hyper});
		\item Compute the $\epsilon$ indicator (Section~\ref{epsilon});
		\item Store each quality indicator result in a separate vector.
	\end{itemize}
\item Algorithms are compared in three ways:
	\begin{itemize}
	\item A group of algorithms is compared using their respective quality indicator's mean value and standard deviation.
	Algorithms with mean value inside the standard deviation of the best mean value are considered equally good.
	\item Verify the statistical significance between a pair of algorithms with a non-parametric Mann-Whitney test \cite{hollander1999nonparametric}. The alternative hypothesis that one method has a better (smaller) quality indicator than the other is accepted if the p-value is lower than $0.05$. 
	\item Calculate the 50\% attainment surface (Section~\ref{surf}) based on the solution sets.
	\end{itemize}
\end{enumerate}


\subsection{Quality Indicators}

In this article, to compare the quality of the algorithms, the hypervolume indicator \cite{zitzler1998multiobjective},\cite{beume2009complexity} and the $\epsilon$~indicator \cite{zitzler2003performance} are used.
These unary quality indicators were recommended by \cite{fonseca2005tutorial}, since they are based on different preference information.
The following subsections define these quality indicators.

\subsubsection{Hypervolume indicator}
\label{hyper}
The hypervolume indicator ($I_h$) is defined as the difference between the hypervolume of the Pareto front and the hypervolume of the non-dominated solution set in objective space \cite{zitzler1998multiobjective},\cite{beume2009complexity}. This indicator requires a reference point for the calculation, therefore the nadir point is used in this article. 

\subsubsection{$\epsilon$ indicator}
\label{epsilon}
The $\epsilon$~indicator ($I_\epsilon$) is defined as the minimum factor $\epsilon$ by which a non-dominated approximation set (i.e., set of objective vectors which do not dominate each other) is worse than the Pareto optimal front. 
Let $a$ and $p$ be vectors in $Z$ (the objective space) with $Z \subseteq \mathbb{R}^{+d}$ where $d$ is the number of objectives, then the $\epsilon$~dominance between two vectors is defined by Equation~\ref{eps_dominance}.
\begin{equation} 
\label{eps_dominance}
	a \succeq_{\epsilon} p \equiv	\forall i\in[1,d] : a_i \leq \epsilon \cdot p_i.
\end{equation}
Then, according to \cite{zitzler2003performance}, the $\epsilon$~indicator is formally defined in Equation~\ref{epsilon_indicator}.
\begin{equation} 
\label{epsilon_indicator}
	I_\epsilon(T) = \inf_{\epsilon\in\mathcal{R}}\{\forall p \in O~  \exists a \in T : a \succeq_{\epsilon} p \},
\end{equation}
where $T$ is the target approximation set and $O$ is the Pareto optimal set. In this paper $O$ refers to a reference set which approximates the Pareto optimal set.

As shown in \cite{okabe2003critical}, quality indicators may be misleading. Therefore, when visually possible, attainment surfaces were also computed for the comparison.

\subsection{Attainment Surfaces}
\label{surf}

Attainment surface (AS) is the boundary in objective space of the dominated area for a single run of an algorithm. They are important because such surfaces show detailed information about the performance differences between algorithms.
To infer a statistically significant attainment surface, multiple runs of the algorithms are required and an approximated mean result is calculated. Usually, the 50\% attainment surface is used as a mean measure approximation, which is defined as the area dominated by at least 50\% of the approximation sets \cite{grunert2001inferential},\cite{fonseca1996performance}.
In this paper, the code provided by \cite{lópez2010exploratory} is used to obtain the 50\% attainment surfaces.

\section{Experiments}
\label{problems}

Some of the usual benchmarks of multi-objective problems poorly represent important classes such as non-separable and multimodal problems.
Therefore, this paper makes use of a relatively recent set of tests called WFG \cite{huband2006review}.
The WFG set of problems present a varied set of properties which can test the scalability of algorithms in both parameters and number of objectives.
In Table~\ref{wfg_properties} there is a summary of the characteristics of its test problems.
The WFG Toolkit makes use of position and distance parameters.
In one hand, when a distance parameter is modified the new solution may dominate, be dominated or be equivalent to the previous one.
On the other hand, when a position parameter is modified the new solution is either incomparable or equivalent to the previous one.
Tests were performed for the WFG problems with $20$ distance parameters and $4$ position parameters, resulting in $24$ parameters to be optimized.

\begin{table*}
\centering
\caption{Properties of the WFG test problems.}
\resizebox{14cm}{!}
{
\begin{tabular}{|c|c|c|c|c|c|}
\hline
	Problem & Obj. & Separable & Modality & Bias & Geometry \\ \hline
	WFG1 & $f_{1:M} $ & yes & uni & polynomial,flat & convex,mixed \\ \hline
	WFG2 & $f_{1:M-1} $ & no & uni & - & convex,disconnected \\ 
	     & $f_{M} $ & no & multi & - & \\ \hline
	WFG3 & $f_{1:M} $ & no & uni & - & linear,degenerate \\ \hline
	WFG4 & $f_{1:M} $ & yes & multi & - & concave \\ \hline
	WFG5 & $f_{1:M} $ & yes & deceptive & - & concave \\ \hline
	WFG6 & $f_{1:M} $ & no & uni & - & concave \\ \hline
	WFG7 & $f_{1:M} $ & yes & uni & parameter dependent & concave \\ \hline
	WFG8 & $f_{1:M} $ & no & uni & parameter dependent & concave \\ \hline
	WFG9 & $f_{1:M} $ & no & multi,deceptive & parameter dependent & concave \\ \hline
\end{tabular}
}
\label{wfg_properties}
\end{table*}


\subsection{Results and Discussions}
\label{experiments}

Each empirical attainment surface and quality indicator was calculated based on $30$ solution sets, which were obtained from multiple independent runs of the algorithm in question.
Different seeds were used for each algorithm run.
Both the maximum number of generations and the \textit{total subpopulation size}\footnote{Note that the variables \textit{subpopulation size} and \textit{total subpopulation size} are different from each other. The total subpopulation is defined in Section~\ref{gsf}.} (or population size in the case of panmictic algorithms) were fixed to respectively $25000$ and $100$.
This fact assures that all algorithms have the same number of evaluations.


\subsection{Choice of Parameters}

Table~\ref{parameters_table} shows the parameters used for GDE3. 
They correspond to the same used by Kukkonen and Lampinen \cite{kukkonen2007performance}.
The reader may observe that when compared with usual single-objective DE's settings, the parameters of all algorithms possess a lower value of $CR$ and~$F$. 
This happens because multi-objective optimization maintain a high diversity. 
Therefore, it is not necessary to have a higher value of $F$ or $CR$ for better exploration of the search space, because individuals are different enough and the trial vectors are also suitably different.
Tests with even smaller values of $F$ were shown to improve the coverage ($F=0.1$), but with great impacts on the distance to the Optimal Pareto Front (OPF). 
The gain in coverage was not enough to surpass SAN's coverage and the distance to the front was poorer enough, such that GDE3 was surpassed by SAN in all problems tested (even on some problems that it performed similarly to SAN with $F=0.5$).

In the case of GSA's algorithms, $F$ should be logically an even lower value.
This is justified by the fact that GSA's subpopulations are usually very different from each other.
We conducted preliminary tests with $F=0.5$ and many results were the same as the ones obtained with $F=0.1$, though some problems showed as expected a slightly worse result.
For MONA and SAN, the novelty parameters were decided upon a quality-efficiency trade-off, with both algorithms having the same fixed parameters.

Regarding the chosen subpopulation sizes of SAN and SAGDE, they are directly related to subpopulation's algorithm strength to ``mix" the solutions of the single-objective DEs' subpopulations. 
Some subpopulations ``mix" better the solutions than others (directly related to the coverage of the OPF), requiring a smaller subpopulation size (MONA subpopulation), while other subpopulations require a bigger subpopulation size to get a similar coverage (GDE3).
This happens specially because GDE3 have various strategies and coverage is just one of its strategies.
Recall that in SAN and SAGDE there are two single-objective DEs. 
These algorithms explore the problems as shown in Figure~\ref{des_only} and discussed in Section~\ref{gsa}.
Therefore, ``mixing" the solution is necessary for coverage and this is only achieved by other subpopulations (GDE3 and MONA subpopulations for respectively SAGDE and SAN).

\begin{table}
\centering
\caption{Parameter's Table. The first two ratios of the ${\cal S}$ vector correspond to the subpopulations of DEs used and the third ratio is either MONA (for the SAN) or GDE3 (for the SAGDE).}
{
\begin{tabular}{|c|c|}
\hline
	 & GDE3 \\ \hline
	$CR$ & 0.1 \\ \hline
	$F$ & 0.5 \\ \hline \hline
	 & MONA \\ \hline
	$CR$ & 0.1 \\ \hline
	$F$ & 0.1 \\ \hline 
	$n_{inc}$ & 1.1 \\ \hline
	$n_{dec}$ & 0.999 \\ \hline
	$n_{a}$ & 1 \\ \hline
	$n_{r}$ & 50000 \\ \hline \hline
	 & SAGDE \\ \hline
	$CR$ & 0.1 \\ \hline
	$F$ & 0.1 \\ \hline 
	${\cal IM}$ & uniform \\ \hline
	${\cal S}$ &$ (0.1,0.1,0.8)$ \\ \hline \hline
	 & SAN \\ \hline
	$CR$ & 0.1 \\ \hline
	$F$ & 0.1 \\ \hline 
	${\cal IM}$ & uniform \\ \hline
	${\cal S}$ & $(0.3,0.3,0.4)$ \\ \hline
	$n_{inc}$ & 1.1 \\ \hline
	$n_{dec}$ & 0.999 \\ \hline
	$n_{a}$ & 1 \\ \hline
	$n_{r}$ & 50000 \\ \hline
\end{tabular}
}
\label{parameters_table}
\end{table}

\subsection{Study on Bi-objective Optimization}
\label{comparison_1}

Tests were performed for the WFG problems with two objectives. 
Parameters used by the algorithms are fixed and summarized in Table~\ref{parameters_table}.

\begin{figure*}[h!]
 \centering
 \includegraphics[scale=0.8]{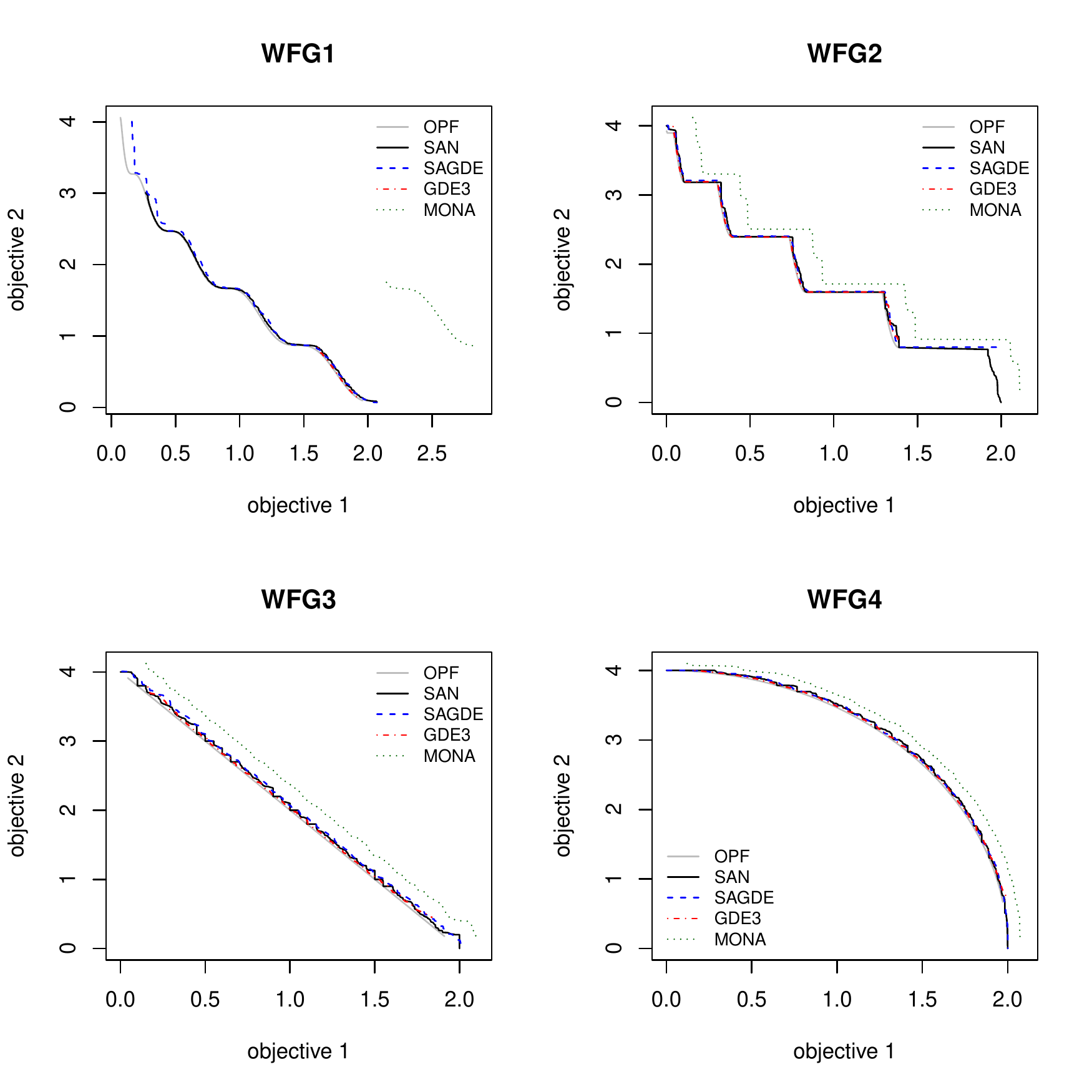}
 \caption{50\% attainment surfaces for the WFG Toolkit problems (minimization problems). Calculated for $30$ independent runs.}
 \label{surfaces1}
\end{figure*}

\begin{figure*}[h!]
 \centering
 \includegraphics[scale=0.8]{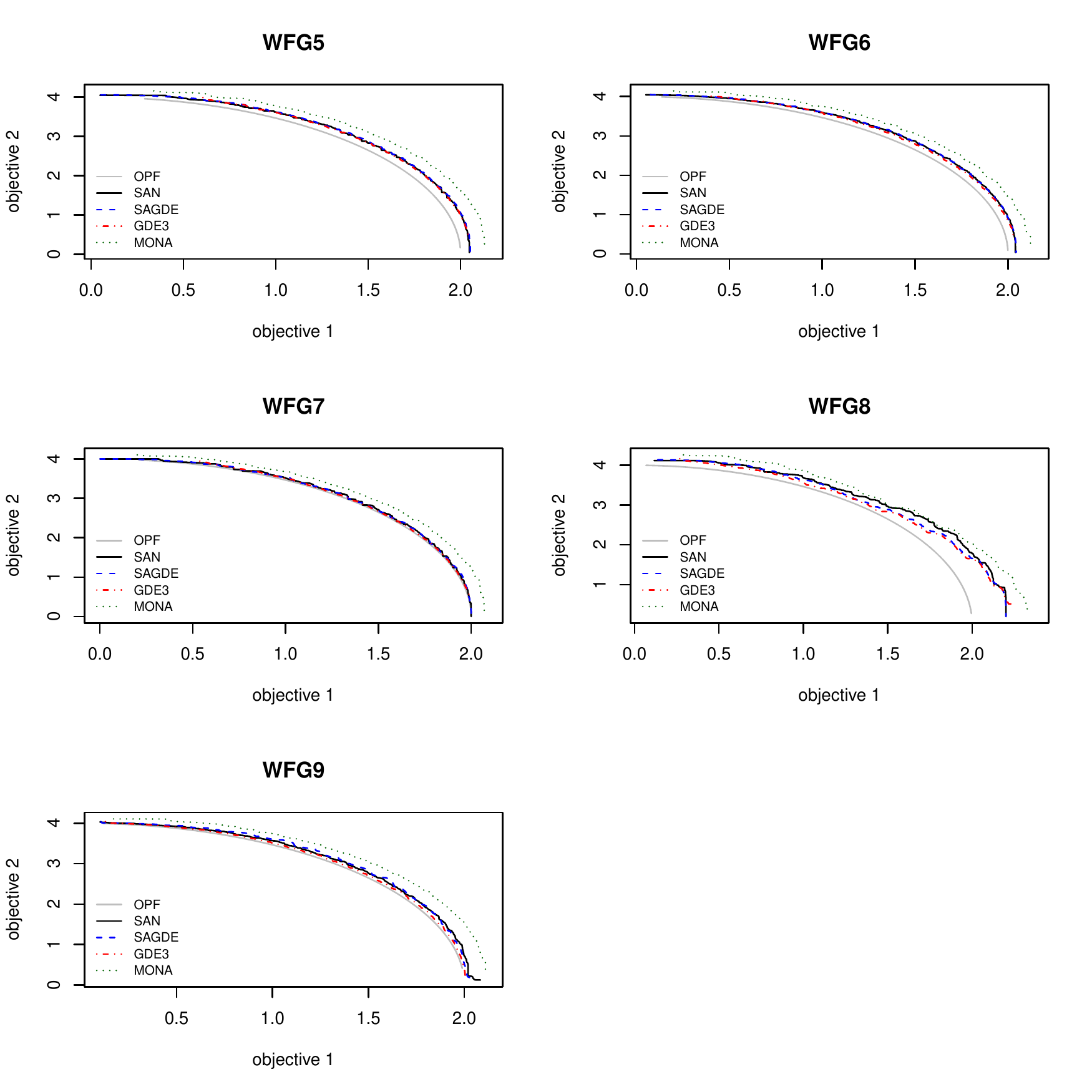}
 \caption{50\% attainment surfaces for the WFG Toolkit problems (minimization problems). Calculated for $30$ independent runs.}
 \label{surfaces2}
\end{figure*}

The comparison between the 50\% attainment surfaces of SAN, SAGDE, GDE3 and MONA is shown in Figures~\ref{surfaces1} and~\ref{surfaces2}.
Before discussing the results it is necessary to shown Tables~\ref{mean_sd_e} and~\ref{mean_sd_h} with the mean and standard deviation (sd) of $\epsilon$ and hypervolume quality indicators as well as Tables~\ref{statistical_e} and~\ref{statistical_h} with the statistical significance of both quality indicators.
Most of the time the tables and figures agree with each other.
Therefore, when not stated otherwise, the discussion concerns the overall behavior of all three comparisons (attainment surfaces, mean/sd and statistical hypothesis testing)
For more information on the construction of these tables and figures please refer to Section~\ref{comparison_methodology} or to the tables and figures themselves.

\begin{table}
\centering
\caption{$\epsilon$ indicator's mean and standard deviation for SAN, SAGDE, GDE3 and MONA. For each problem the best mean value as well as the other mean values inside the standard variation of the best mean value are marked in \textbf{bold}.}
\begin{tabular}{|l|l|l|l|l|}
\hline
& SAN & SAGDE & GDE3 & MONA \\ \hline
 Problems & mean (sd) & mean (sd) & mean (sd) & mean (sd) \\ \hline
     WFG1 &  $\mathbf{0.20 (0.03)}$ & $\mathbf{0.18 (0.09)}$ &  $1.53 (0.02)$& $2.06 (0.06)$\\
     WFG2 &  $\mathbf{0.05 (0.01)}$ & $0.10 (0.11)$&  $0.42 (0.37)$& $0.18 (0.02) $ \\
     WFG3 &  $\mathbf{0.07 (0.01)}$ & $0.14 (0.03)$&  $0.29 (0.13)$& $0.19 (0.02) $\\
     WFG4 &  $\mathbf{0.07 (0.01)}$ & $0.10 (0.03)$&  $0.28 (0.23)$& $0.14 (0.01) $\\
     WFG5 &  $\mathbf{0.12 (0.01)}$ & $0.17 (0.03)$&  $0.37 (0.10)$& $0.22 (0.02) $\\
     WFG6 &  $\mathbf{0.11 (0.01)}$ & $0.16 (0.04)$&  $0.44 (0.30)$& $0.19 (0.02) $\\
     WFG7 &  $\mathbf{0.08 (0.01)}$ & $0.10 (0.02)$&  $0.41 (0.16)$& $0.16 (0.01) $\\
     WFG8 &  $\mathbf{0.23 (0.01)}$ & $0.29 (0.10)$&  $0.41 (0.31)$& $0.33 (0.02)$ \\
 WFG9 &  $\mathbf{0.09 (0.01)}$ & $0.15 (0.06)$&  $0.15 (0.18)$& $0.18 (0.01) $\\
\hline
\end{tabular}
\label{mean_sd_e}
\end{table}

\begin{table}
\centering
\caption{Hypervolume indicator's mean and standard deviation for SAN, SAGDE, GDE3 and MONA. For each problem the best mean value as well as the other mean values inside the standard variation of the best mean value are marked in \textbf{bold}.}
\begin{tabular}{|l|l|l|l|l|}
\hline
& SAN & SAGDE & GDE3 & MONA \\ \hline
 Problems &  mean (sd) & mean (sd) & mean (sd) & mean (sd) \\ \hline
      WFG1&   $\mathbf{0.21 (0.04)}$ &  $\mathbf{0.24 (0.12)}$& $3.37 (0.08)$ & $6.23 (0.15)$ \\
      WFG2&   $\mathbf{0.02 (0.01)}$ &  $0.10 (0.10)$& $0.25 (0.20)$ & $0.67 (0.11)$ \\ 
      WFG3&   $\mathbf{0.13 (0.02)}$ &  $0.23 (0.04)$& $\mathbf{0.13 (0.02)}$ & $0.75 (0.15)$ \\
      WFG4&   $\mathbf{0.09 (0.01)}$ &  $0.11 (0.01)$& $\mathbf{0.08 (0.01)}$ & $0.46 (0.05)$ \\
      WFG5&   $\mathbf{0.37 (0.02)}$ &  $0.42 (0.02)$& $\mathbf{0.35 (0.02)}$ & $0.86 (0.10)$ \\
      WFG6&   $\mathbf{0.34 (0.02)}$ &  $\mathbf{0.37 (0.03)}$& $\mathbf{0.32 (0.23)}$ & $0.78 (0.12)$ \\
      WFG7&   $\mathbf{0.10 (0.01)}$ &  $\mathbf{0.12 (0.01)}$& $\mathbf{0.10 (0.03)}$ & $0.51 (0.07)$ \\
      WFG8&   $0.87 (0.05)$ &  $0.88 (0.24)$& $\mathbf{0.62 (0.08)}$ & $1.28 (0.10)$ \\
      WFG9&   $\mathbf{0.21 (0.02)}$ &  $0.35 (0.20)$& $\mathbf{0.15 (0.12)}$ & $0.72 (0.03)$ \\
\hline
\end{tabular}
\label{mean_sd_h}
\end{table}

\begin{table}
\centering
\caption{P-values of comparison between SAN, SAGDE, GDE3 and MONA algorithms with Mann-Whitney significance test using the $\epsilon$ indicator. 
Results are marked in \textbf{bold} when the null hypothesis is rejected with a significance level of $\alpha=0.05$.
The alternative hypothesis is that the algorithm in the row is statistically better (smaller quality indicator) than the algorithm in the column.}
\begin{tabular}{|l|l|l|l|l|l|}
\hline
Algorithm & Problem  & SAN & SAGDE & GDE3 & MONA\\ \hline
 \multirow{9}{*}{SAN} & WFG1  &  & $0.99$ & $ \mathbf{8.4e-18}$ & $ \mathbf{8.4e-18}$ \\
        & WFG2& $  $ & $ \mathbf{0.01}$ & $ \mathbf{1.7e-4}$ & $ \mathbf{8.4e-18}$ \\
        & WFG3& $  $ & $ \mathbf{4.4e-11}$ & $ \mathbf{8.4e-18}$ & $\mathbf{8.4e-18} $ \\
        & WFG4& $  $ & $ \mathbf{4.1e-8}$ & $ \mathbf{1.2e-11}$ & $\mathbf{8.4e-18}$ \\
        & WFG5& $  $ & $ \mathbf{3.9e-11} $ & $ \mathbf{8.4e-18}$ & $\mathbf{8.4e-18}$ \\
	& WFG6& $  $ & $ \mathbf{1.5e-10} $ & $ \mathbf{8.4e-18}$ & $\mathbf{8.4e-18}$ \\
	& WFG7& $  $ & $ \mathbf{2.1e-7}$ & $ \mathbf{8.4e-18}$ & $\mathbf{8.4e-18}$ \\
	& WFG8& $  $ & $ 0.35$ & $ \mathbf{2.1e-4}$ & $\mathbf{8.4e-18}$ \\
	& WFG9& $  $ & $ \mathbf{1.3e-6}$ & $ \mathbf{5.4e-4}$ & $\mathbf{8.4e-18}$ \\ \hline
 \multirow{9}{*}{SAGDE}	& WFG1 &  $\mathbf{5.0e-3}$ & & $\mathbf{8.4e-18}$  & $\mathbf{8.4e-18}$ \\
        & WFG2& $ 0.98  $ && $\mathbf{2.6e-3}$ & $\mathbf{1.3e-10}$ \\
	& WFG3& $ 0.99 $ && $ \mathbf{1.2e-7}$  & $\mathbf{9.3e-8} $ \\
	& WFG4& $ 0.99 $ && $ \mathbf{9.9e-5}$ & $\mathbf{2.4e-6}$ \\
	& WFG5& $ 0.99 $ && $ \mathbf{4.8e-14}$ & $\mathbf{9.4e-8}$ \\
	& WFG6& $ 0.99 $ && $ \mathbf{8.3e-13}$ & $\mathbf{1.2e-4}$ \\
	& WFG7& $ 0.99 $ && $ \mathbf{5.6e-16}$ & $\mathbf{9.1e-12}$ \\
	& WFG8& $ 0.65 $ && $ \mathbf{0.02}$ & $\mathbf{2.3e-3}$ \\
	& WFG9& $ 0.99 $ && $ 0.95$  & $\mathbf{0.01}$ \\ \hline
 \multirow{9}{*}{GDE3}	& WFG1 & $ 1$ & $1$ & $ $ & $ \mathbf{8.4e-18}$ \\
	& WFG2 & $ 0.99$ & $ 0.99 $ & $ $ & $ 0.63 $ \\
	& WFG3 & $ 1$    & $ 0.99 $ & $ $ & $ 0.99 $ \\
	& WFG4 & $ 0.99$ & $ 0.99 $ & $ $ & $ 0.92$ \\
	& WFG5 & $ 1$    & $ 0.99 $ & $ $ & $ 0.99$ \\
	& WFG6 & $ 1$    & $ 0.99 $ & $ $ & $ 0.99$ \\
	& WFG7 & $ 1$    & $ 1   $  & $ $ & $ 0.99$ \\
	& WFG8 & $ 0.99$ & $ 0.97 $ & $ $ & $ 0.12$ \\
	& WFG9 & $ 0.99$ & $ \mathbf{0.04} $ & $ $ & $ \mathbf{5.4e-7}$ \\ \hline
 \multirow{9}{*}{MONA}	& WFG1 & $ 1 $ & $1$ & $1$ &  \\
        & WFG2& $ 1 $ & $ 0.99$ & $0.37$  &  \\
	       & WFG3& $ 1 $ & $ 0.99$ & $\mathbf{4.1e-4}$ & \\
	& WFG4& $ 1 $ & $ 0.99$ & $0.07$ & \\
	& WFG5& $ 1 $ & $ 0.99$ & $\mathbf{5.4e-10}$ & \\
	& WFG6& $ 1 $ & $ 0.99$ & $\mathbf{1.7e-10}$ & \\
	& WFG7& $ 1 $ & $ 0.99$ & $\mathbf{1.8e-12}$ & \\
	& WFG8& $ 1 $ & $ 0.99$ & $0.87$ & \\
	& WFG9& $ 1 $ & $ 0.98$ & $0.99$ & \\
\hline
\end{tabular}
\label{statistical_e}
\end{table}

\begin{table}
\centering
\caption{P-values of comparison between SAN, SAGDE, GDE3 and MONA algorithms with Mann-Whitney significance test using the hypervolume indicator. 
Results are marked in \textbf{bold} when the null hypothesis is rejected with a significance level of $\alpha=0.05$.
The alternative hypothesis is that the algorithm in the row is statistically better (smaller quality indicator) than the algorithm in the column.}
\begin{tabular}{|l|l|l|l|l|l|}
\hline
Problem & Algorithm  & SAN & SAGDE & GDE3 & MONA\\ \hline
 \multirow{9}{*}{SAN}	& WFG1 & $  $ & $ 0.40$ & $ \mathbf{8,4e-18}$ & $ \mathbf{8.4e-18}$ \\
        & WFG2 & $  $ & $ \mathbf{3.7e-13}$ & $ \mathbf{7.6e-12}$ & $ \mathbf{8.4e-18}$ \\
        & WFG3 & $  $ & $ \mathbf{7.8e-14}$ & $0.53$ & $ \mathbf{8.4e-18}$ \\
      	& WFG4 & $  $ & $ \mathbf{1.6e-4}$ & $ 0.99$ & $ \mathbf{8.4e-18}$ \\
     	& WFG5 & $  $ & $ \mathbf{4.5e-12}$ & $ 0.99$ & $ \mathbf{8.4e-18}$ \\
        & WFG6 & $  $ & $ \mathbf{1.6e-4}$ & $ 0.99$ & $ \mathbf{8.4e-18}$ \\
        & WFG7 & $  $ & $ \mathbf{6.8e-4}$ & $ 0.98$ & $ \mathbf{8.4e-18}$ \\
	& WFG8 & $  $ & $ 0.74$ & $ 0.99$ & $ \mathbf{8.4e-18}$ \\
	& WFG9 & $  $ & $ 0.46$ & $ 0.99$ & $ \mathbf{8.4e-18}$ \\\hline
 \multirow{9}{*}{SAGDE}	& WFG1 & $0.60$ &  & $\mathbf{8.4e-18}$ & $ \mathbf{8.4e-18}$ \\
        & WFG2& $ 0.99  $ &  & $0.1$  & $ \mathbf{3.1e-15}$ \\
	& WFG3& $ 0.99 $ &  & $ 0.99$ & $\mathbf{8.4e-18} $ \\
	& WFG4& $ 0.99 $ &  & $ 0.99$ & $\mathbf{8.4e-18}$ \\
	& WFG5& $ 0.99 $ &  & $ 1$   & $\mathbf{8.4e-18}$ \\
	& WFG6& $ 0.99 $ &  & $ 0.99$ & $\mathbf{8.4e-18}$ \\
	& WFG7& $ 0.99 $ &  & $ 0.99$ & $\mathbf{8.4e-18}$ \\
	& WFG8& $ 0.26 $ &  & $ 0.99$& $\mathbf{1.2e-9}$ \\
	& WFG9& $ 0.53 $ &  & $ 0.99$& $\mathbf{8.4e-18}$ \\ \hline
 \multirow{9}{*}{GDE3}	& WFG1 & $ 1 $ & $ 1 $ & $ $ & $ \mathbf{8.4e-18}$ \\
	& WFG2 & $ 0.99   $ & $ 0.89 $ & $ $   & $ \mathbf{2.5e-16}$  \\
	& WFG3 & $ 0.46   $ & $ \mathbf{2.9e-14}$ & $ $ & $\mathbf{8.4e-18} $ \\
	& WFG4 & $\mathbf{1.6e-4} $ & $ \mathbf{9.3e-10}$ & $ $ & $\mathbf{8.4e-18}$  \\
	& WFG5 & $\mathbf{5.1e-4} $ & $ \mathbf{1.0e-16}$ & $ $ & $\mathbf{8.4e-18}$  \\
	& WFG6 & $ \mathbf{2.6e-8}$ & $ \mathbf{3.3e-11}$ & $ $ & $\mathbf{2.4e-13}$  \\
	& WFG7 & $ \mathbf{0.01}   $ & $ \mathbf{4.6e-4}$ & $ $  & $\mathbf{8.4e-18}$  \\
	& WFG8 & $ \mathbf{1.9e-13}$ & $ \mathbf{1.0e-9}$ & $ $  & $\mathbf{8.4e-18}$  \\
	& WFG9 & $ \mathbf{7.1e-10}$ & $ \mathbf{2.3e-10}$ & $ $ & $\mathbf{5.9e-17}$  \\ \hline
 \multirow{9}{*}{MONA}	& WFG1 & $ 1 $ & $ 1 $ & $ 1$ & $1$ \\
        & WFG2& $ 1 $ & $ 0.99$ & $1 $ & \\
	& WFG3& $ 1 $ & $ 1 $   & $1 $ &  \\
	& WFG4& $ 1 $ & $ 1 $   & $1 $ & \\
	& WFG5& $ 1 $ & $ 1 $   & $1 $ & \\
	& WFG6& $ 1 $ & $ 1 $   & $0.99 $ & \\
	& WFG7& $ 1 $ & $ 1 $   & $1 $ & \\
	& WFG8& $ 1 $ & $ 0.99$ & $1 $ &  \\
	& WFG9& $ 1 $ & $ 1 $   & $1 $ &  \\
\hline
\end{tabular}
\label{statistical_h}
\end{table}


Regarding the comparison between SAGDE and GDE3. 
SAGDE is significantly better than the GDE3 in the WFG1 for both quality indicators (clearly observable in Tables~\ref{statistical_e} and~\ref{statistical_h} but also present in the other tables and figures). 
However, the quality indicators do not agree in the remaining problems, which suggests that there is just a trade-off but not an explicit advantage in these problems.
SAGDE tends to achieve a better coverage of the OPF, while GDE3 is closer to the OPF albeit having a slightly poorer coverage of the front.
Consequently, depending on whether coverage or proximity to the front is more important, the algorithm designer may choose one or the other algorithm.



MONA achieved poor outcomes on all problems against all algorithms. 
Maybe the exceptions are the better coverage when compared against GDE3 in WFG3, WFG5, WFG6 and WFG7 problems (see Table~\ref{statistical_e} and Table~\ref{mean_sd_e}).  
Even so, the combined subpopulations of MONA and the single-objective DEs in the SAN obtained state of art quality.
Notice also that inside the SAGDE and SAN there is respectivelly a GDE3 and a MONA subpopulation.
The GDE3 subpopulation inside SAGDE is bigger than the MONA subpopulation inside SAN, however, GDE3 subpopulation still ``mix" the solutions worse than MONA (resulting in poorer coverage).
Demonstrating MONA's good ability of expanding and mixing results.

\begin{figure}[h!]
 \centering
 \includegraphics[scale=0.8]{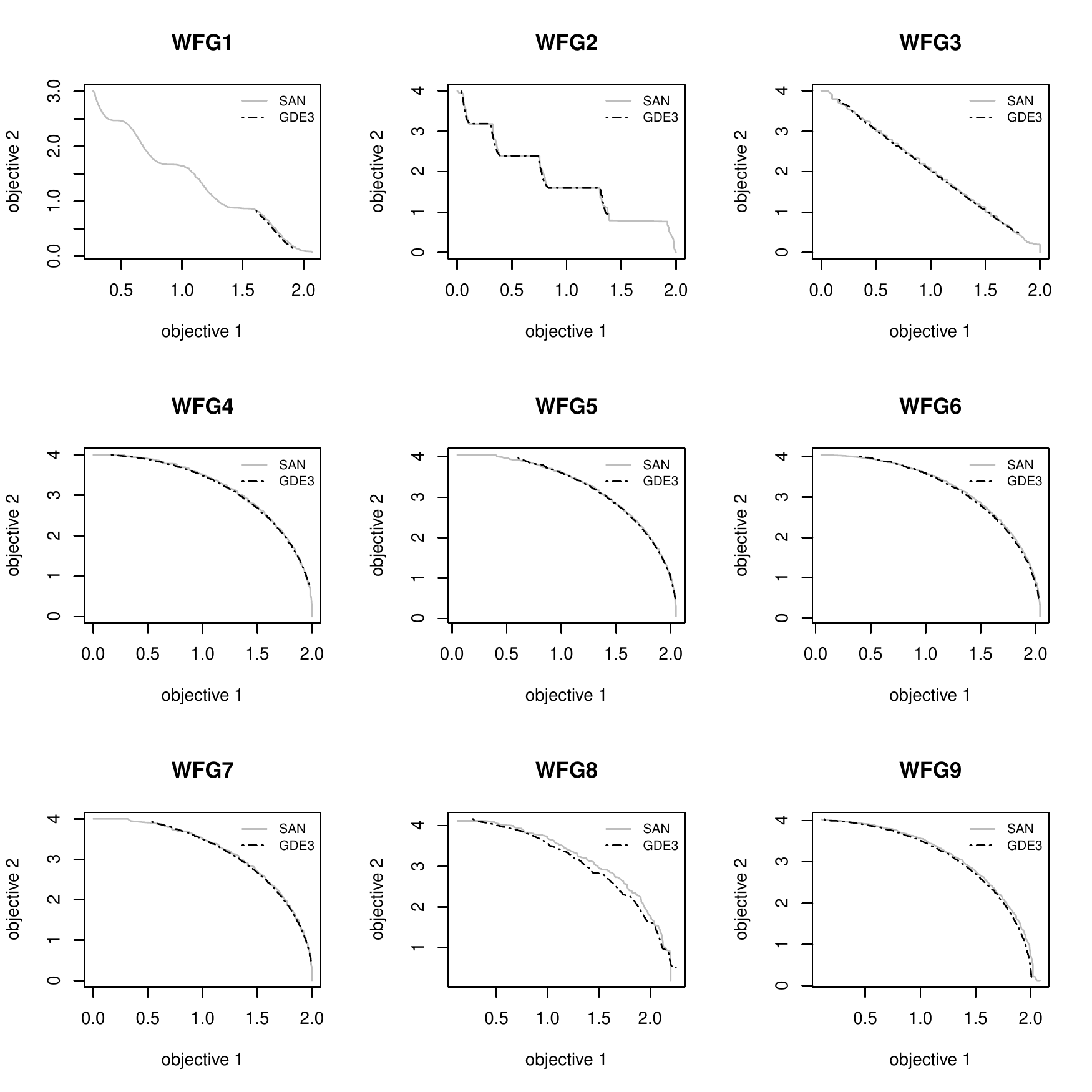}
 \caption{50\% attainment surfaces for the WFG Toolkit problems (minimization problems). Calculated for $30$ independent runs.}
 \label{surfaces3}
\end{figure}

\begin{figure}[h!]
 \centering
 \includegraphics[scale=0.6]{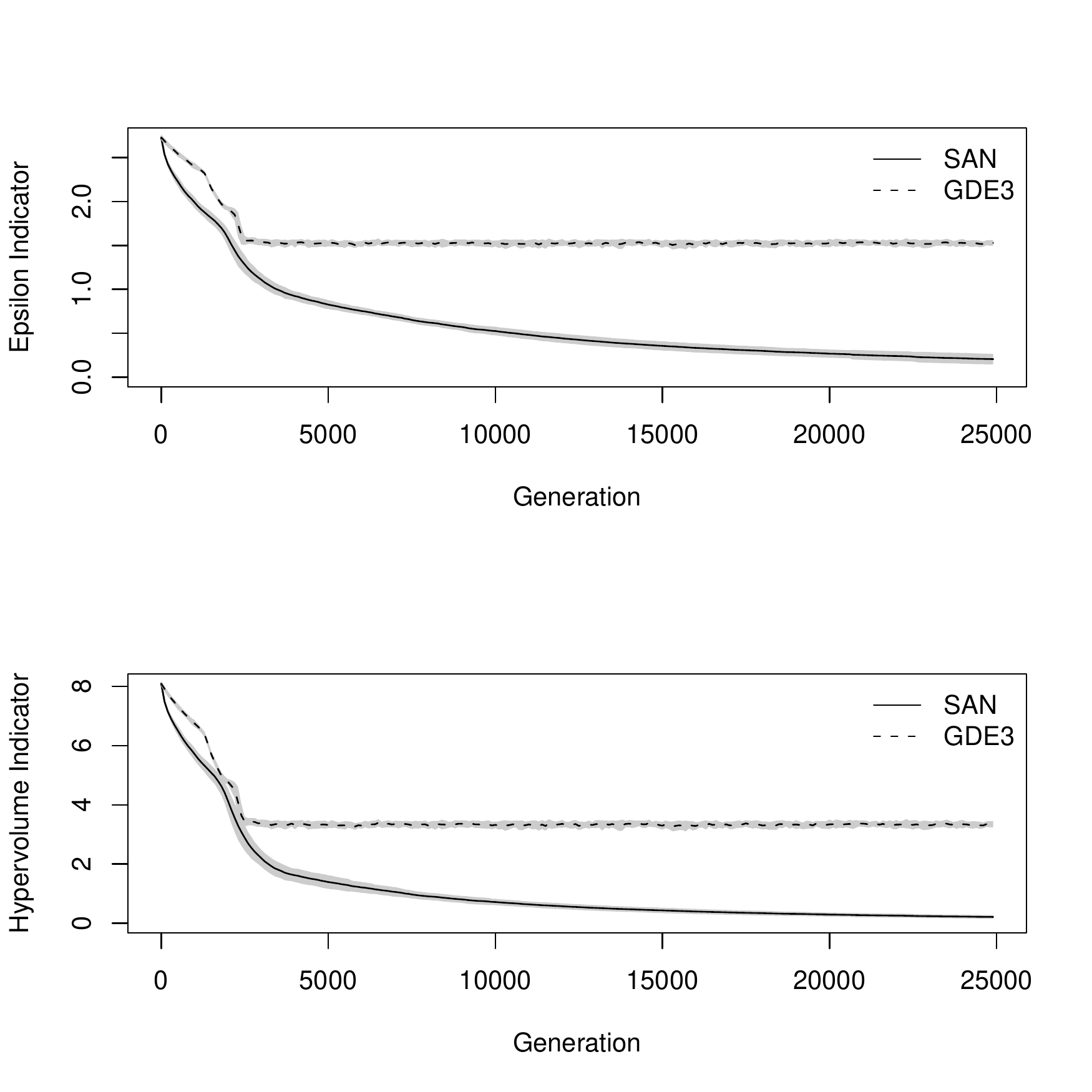}
 \caption{Hypervolume and $\epsilon$ indicators throughout the generations for both SAN and GDE3 algorithms in the WFG1 problem (the confidence interval of one standard deviation away from the mean is shown in grey). The curve was averaged over $30$ independent runs.}
 \label{progress}
\end{figure}

Concerning the comparison of both GSAs, the experiments demonstrate a surprisingly better overall result of the SAN over the SAGDE, as the SAN is simpler and based on the MONA, an algorithm which achieved poor results on all tests. 
This fact might seem surprising at first glance, but looking from a different point of view, it is possible to understand those results if we take into account the GSF's structure. 
Recall that the more different two strategies are, the more the subpopulation's framework benefits from it. 
This happens because a similar strategy will also produce similar individuals in different subpopulations, using more resources for less exploration and diversity.

The comparison between SAN and GDE3 is a bit more complicated.
First of all, Tables~\ref{statistical_e} and~\ref{statistical_h} show that SAN outperforms GDE3 according to both quality indicators in WFG1, WFG2 and WFG3 while the remaining problems have contrasting results of $\epsilon$ and hypervolume indicators.
Consequently, GDE3 is comparable with SAN only in the concave problems, which have easier shapes of Pareto front. 
However, the statistical hypothesis testing does not tell us by how much is the difference (it only tells if it is bigger or not with some significance).
Table~\ref{mean_sd_e} shows unsurprisingly that SAN outperforms GDE3 by a great difference in respect to the $\epsilon$ indicator.
But according to Table~\ref{mean_sd_h}, in all problems where GDE3 surpassed SAN statistically, GDE3 is shown to be close (inside GDE3's standard deviation) to the SAN in all problems but WFG8 (the reason why this happens is show on Section~\ref{empirical_forces}, where it is demonstrated that both algorithms have not converged yet in WFG8).
In fact, to the knowledge of the authors, SAN achieved the best performance to date over all of the WFG's problems with two objectives.
Additionally, for a clearer analysis, Figure~\ref{surfaces3} shows only SAN and GDE3 attainment surfaces and Figure~\ref{progress} delineates the behavior of the quality indicators throughout the evolution of problem WFG1.
It can be seen that both SAN and GDE3 start with similar quality (i.e., no initialization difference).
However, SAN is always superior to GDE3 in quality soon after the starting point. 







Both of the subpopulation algorithms presented here are strongly based on the division of the related panmictic algorithm's strategies into different subpopulations and with the results showing strong benefits of the subpopulation algorithms over the panmictic ones. 
This raises the question of weather the competition between different strategies inside one population can have deleterious consequences for an algorithm.
In fact, SAN, which has entirely different subpopulations in terms of objectives (small or inexistent conflict inside the same subpopulation), was able to achieve the best results. 
Contrast this with the GDE3, which has three conflicting objectives inside its panmictic population (the two objectives of the problem and the diversity objective).	
Section~\ref{explanation} will touch this hypothesis more extensively and with an experimental test.


\subsection{Study on Many-objective Optimization}
\label{comparison_2}

In this study, we increased the number of objectives to five. Aside from that, the same WFG problems were used and all other problem's parameters were kept as before.
Most of the algorithms' parameters remained the same as well with the only exception being vector ${\cal S}$, which depends on the number of objectives. 
The new set of parameters is shown in Table~\ref{parameters_table2}.

\begin{table}
\centering
\caption{Parameter's Table. The first five ratios of the ${\cal S}$ vector correspond to the subpopulations of the DEs used and the last ratio corresponds to the MONA.}
\begin{tabular}{|c|c|}
\hline
	 & GDE3 \\ \hline
	$CR$ & 0.1 \\ \hline
	$F$ & 0.5 \\ \hline \hline
	 & SAN \\ \hline
	$CR$ & 0.1 \\ \hline
	$F$ & 0.1 \\ \hline 
	${\cal IM}$ & uniform \\ \hline
	${\cal S}$ & $(0.1,0.1,0.1,0.1,0.1,0.5)$ \\ \hline
	$n_{inc}$ & 1.1 \\ \hline
	$n_{dec}$ & 0.999 \\ \hline
	$n_{a}$ & 1 \\ \hline
	$n_{r}$ & 50000 \\ \hline
\end{tabular}
\label{parameters_table2}
\end{table}

\begin{table}
\centering
\caption{$\epsilon$ indicator's mean and standard deviation for SAN and GDE3. For each problem the best mean value as well as the other mean values inside the standard variation of the best mean value are marked in \textbf{bold}.}
\begin{tabular}{|l|c|c|}
\hline
& SAN & GDE3 \\ \hline
 Problems &  mean (sd) & mean (sd) \\ \hline
      WFG1&   $\mathbf{0.55 (0.06)}$  &  $0.97 (0.10) $ \\
      WFG2&   $\mathbf{0.68 (0.11)}$  &  $0.99 (0.31) $ \\ 
      WFG3&   $\mathbf{0.40 (0.07)}$  &  $0.61 (0.10) $ \\
      WFG4&   $\mathbf{0.91 (0.04)}$  &  $1.47 (0.23) $ \\
      WFG5&   $\mathbf{1.27 (0.11)}$  &  $1.67 (0.23)$ \\
      WFG6&   $\mathbf{1.03 (0.05)}$  &  $1.78 (0.34)$ \\
      WFG7&   $\mathbf{0.94 (0.04)}$  &  $1.75 (0.20)$ \\
      WFG8&   $\mathbf{1.13 (0.06)}$  &  $1.56 (0.18)$ \\
      WFG9&   $\mathbf{0.94 (0.06)}$  &  $1.55 (0.20)$ \\
\hline
\end{tabular}
\label{mean_sd_e_many}
\end{table}

\begin{table}
\centering
\caption{Hypervolume indicator's mean and standard deviation for SAN and GDE3. For each problem the best mean value as well as the other mean values inside the standard variation of the best mean value are marked in \textbf{bold}.}
\begin{tabular}{|l|c|c|}
\hline
& SAN & GDE3 \\ \hline
 Problems &  mean (sd) & mean (sd) \\ \hline
      WFG1&   $\mathbf{62.89 (13)}$  &  $997.9 (98) $ \\
      WFG2&   $\mathbf{30.78 (8)}  $  &  $95.5  (40) $ \\ 
      WFG3&   $\mathbf{-92.1 (36)} $  &  $79.3  (30) $ \\
      WFG4&   $\mathbf{-40.7 (48)} $  &  $378.1 (67) $ \\
      WFG5&   $\mathbf{897.8 (173)}$  &  $1088 (53)$ \\
      WFG6&   $\mathbf{636.8 (62)} $  &  $1050 (66)$ \\
      WFG7&   $\mathbf{448.2 (62)} $  &  $1911 (173)$ \\
      WFG8&   $\mathbf{1399 (91)}$  &  $\mathbf{1353 (59)}$ \\
      WFG9&   $\mathbf{-268.0 (71)}$  &  $526.1 (201)$ \\
\hline
\end{tabular}
\label{mean_sd_h_many}
\end{table}

\begin{table}
\centering
\caption{P-values of comparison between SAN and GDE3 algorithms in many-objective problems with Mann-Whitney significance test using $\epsilon$ and hypervolume indicators. 
Results are marked in \textbf{bold} when the null hypothesis is rejected with a significance level of $\alpha=0.05$.
The alternative hypothesis is that the algorithm in the row is statistically better (smaller quality indicator) than the algorithm in the column.}
\begin{tabular}{|l|l|c|c|c|c|}
\hline
& & \multicolumn{2}{c|}{SAN} & \multicolumn{2}{c|}{GDE3} \\
	Algorithm & Problem  & $\epsilon$ & hypervolume   & $\epsilon$ & hypervolume \\ \hline
 \multirow{9}{*}{SAN} & WFG1 & & & $\mathbf{1.0e-16}$ & $\mathbf{8.4e-18}$\\
        & WFG2&  & &$ \mathbf{4.7e-6}$ & $ \mathbf{1.6e-15}$   \\
        & WFG3&  & &$ \mathbf{1.8e-12}$& $ \mathbf{8.4e-18}$   \\
        & WFG4&  & &$ \mathbf{8.4e-18}$& $ \mathbf{8.4e-18} $   \\
        & WFG5&  & &$ \mathbf{2.9e-13}$& $ \mathbf{3.0e-7}$     \\
	& WFG6&  & &$ \mathbf{8.4e-18}$& $ \mathbf{8.4e-18} $   \\
	& WFG7&  & &$ \mathbf{8.4e-18}$& $ \mathbf{8.4e-18} $   \\
	& WFG8&  & &$ \mathbf{3.3e-17}$& $ 0.99 $              \\
	& WFG9&  & &$ \mathbf{8.4e-18}$& $ \mathbf{8.4e-18} $  \\ \hline
 \multirow{9}{*}{GDE3}	& WFG1  &$1$ & $1$ && \\
	& WFG2 & $ 0.99 $ &$ 0.99 $	     &&      \\
	& WFG3 & $ 0.99 $ &$ 1 $	     &&         \\
	& WFG4 & $ 1 $    &$ 1 $	     &&         \\
	& WFG5 & $ 1 $    &$ 1 $	     &&         \\
	& WFG6 & $ 1 $    &$ 1 $	     &&         \\
	& WFG7 & $ 1 $    &$ 1 $	     &&         \\
	& WFG8 & $ 1 $    &$ \mathbf{0.006}$ &&     \\
	& WFG9 & $ 1 $    &$ 1 $     	     &&      \\
\hline
\end{tabular}
\label{statistical_e_many}
\end{table}

Tests with many-objective problems were realized using the SAN, the most prominent algorithm in the bi-objective study from Section~\ref{comparison_1}, and a reference from the state of the art, GDE3.

Tables~\ref{mean_sd_e_many} and~\ref{mean_sd_h_many} display the mean and standard deviation values, while Table~\ref{statistical_e_many} shows the statistical results of the comparison.
SAN is able to converge better in all problems according to both quality indicators except WFG8, where the quality indicators differed in the results (even so, WFG8's hypervolume indicator mean values of SAN and GDE3 are close to each other)
Moreover, in all other problems SAN had very small p-values.
The negative values of the hypervolume indicator means that the samples acquired from the Pareto optimum front dominate a hypervolume smaller than the SAN's dominated hypervolume.
This result may be related to the number and distribution of samples in the OPF generated by the WFG toolkit. 
The same OPF's samples were used to compare both GDE3 and SAN and therefore there is not any bias in the comparison (i.e., GDE3 could have had negative hypervolume as well).

This suggests that SAN should achieve better results when problems increase in complexity. 
Recall that on bi-objective problems, GDE3 was shown to be comparable with SAN only when concave Pareto fronts were present.
Naturally, with the increase in the number of functions to be optimized, the number of conflicting objectives inside panmictic algorithms is expected to increase as well.
This explains the better overall solutions of SAN in all the many-objective problems with many different properties (see Table~\ref{wfg_properties}).

\begin{table}
\centering
\caption{Comparison of the SAN and GDE3 algorithms with Mann-Whitney significance test in many-objective problems. The respective meanings of $\Uparrow$, $\downarrow$ and $\approx$ is that SAN is statistically better, worse or equal to the GDE3.}
\begin{tabular}{|l|ll|}
\hline
 & \multicolumn{2}{c|}{SAN vs GDE3 (many-objective)} \\
	 Problems & $I_\epsilon$(p-value) & $I_h$(p-value) \\ \hline
	WFG1 & $\Uparrow (2.029e-16)$ & $\Uparrow (1.691e-17)$ \\
	WFG2 & $\Uparrow (9.415e-06)$ & $\Uparrow (3.297e-15)$ \\
	WFG3 & $\Uparrow (3.631e-12)$ & $\Uparrow (1.691e-17)$ \\
	WFG4 & $\Uparrow (1.691e-17)$ & $\Uparrow (1.691e-17)$ \\
	WFG5 & $\Uparrow (1.691e-17)$ & $\Uparrow (1.691e-17)$ \\
	WFG6 & $\Uparrow (1.691e-17)$ & $\Uparrow (1.691e-17)$ \\
	WFG7 & $\Uparrow (1.691e-17)$ & $\Uparrow (1.691e-17)$ \\
	WFG8 & $\Uparrow (6.764e-17)$ & $\downarrow (0.013)$ \\
	WFG9 & $\Uparrow (1.691e-17)$ & $\Uparrow (1.691e-17)$ \\
\hline
\end{tabular}
\label{gsan_gde3_many}
\end{table}

\subsection{Explanation}
\label{explanation}

It has been argued before that the algorithms based on the GSF achieve better results since they divide different strategies (algorithms) in distinct populations which avoid both the undesirable conflicts and the prevalence of one strategy over another.
Here, we will present an detailed justification. 

Consider a bi-objective optimization problem being solved with SAN and GDE3. 
For this problem, SAN may be divided into three strategies (i.e., $|A|=3$): one single-objective DE for each of the two objectives and MONA. 
GDE3 has one strategy which is composed of two steps: first selecting individuals based on Pareto dominance (main strategy) and second pruning the population based on a diversity measure (secondary strategy). 

\begin{figure}[h!]
 \centering
 \includegraphics[scale=0.4]{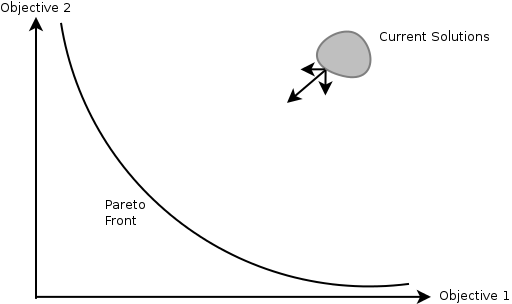}
 \caption{Diagram of the GDE3 with its strategy exposed explicitly as three components of a force. The length of the arrow is related with its intensity.}
 \label{gde3_force}
\end{figure}

\begin{figure}[h!]
 \centering
 \includegraphics[scale=0.4]{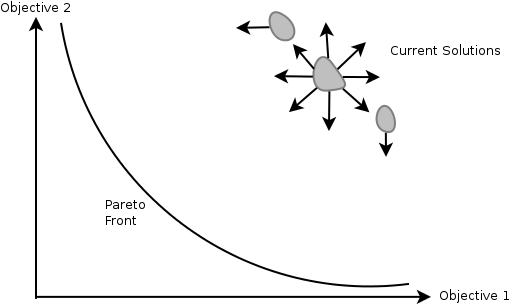}
 \caption{Diagram of the SAN with its strategies exposed explicitly as forces. The single-objective DE forces (dashed gray lines) are perpendicular to each other and the MONA force (circular dashed-point gray line) is a field-force which is stronger with the increase of the distance from the previous individuals.}
 \label{san_force}
\end{figure}

If we see the strategies as a collection of forces capable of changing the positions of solutions, it is possible to draw the most salient force vectors produced by SAN and GDE3 (Figures~\ref{gde3_force} and~\ref{san_force}).
Therefore, for GDE3, the main force points directly to the Pareto front with secondary forces pointing sideways (caused by the pruning strategy). 
In SAN, the single-objective DE's subpopulations' forces point directly to their respective objective's coordinate while the MONA's subpopulation points away from the previous individuals which corresponds approximately to vectors pointing in all directions with the same strength.

This analysis reveals the main problem with GDE3: its forces responsible for spreading are relatively weak. 
The first consequence is, for example, when the problem has a disconnected geometry or bias, the solutions may spread only over a small subset of the optimal front (see problems WFG1, WFG2 of Figures~\ref{surfaces1} and~\ref{surfaces2} or Figure~\ref{surfaces3}).  
Another consequence is that the necessary forces for the solution of problems depends naturally on the problems themselves and if a given problem needs more spreading forces, GDE3 presents many difficulties to spread the solutions. 
For example, over all the WFG's datasets the GDE3 covered poorly the extremes of the Pareto front (see Figures~\ref{surfaces1} and~\ref{surfaces2} or Figure~\ref{surfaces3}) and in the case of many-objective problems, where the Pareto front becomes wider as it expands along various dimensions, it achieved poor results in all tests for both quality indicators (see Table~\ref{gsan_gde3_many}).

Notice that the vectors of the GDE3 are a consequence of its panmictic design which causes inevitably one force to be stronger or weaker relative to the others.
That is, this analysis is inherently connected with the conflicting strategies of panmictic algorithms.


\subsection{Empirical Evaluation of Forces}
\label{empirical_forces}

Measuring the forces empirically can be done in various ways.
If the average solution of each subpopulation in objective space is considered, it is possible to analyze the subpopulation forces throughout the evolution. 
However, the comparison with single population algorithms may be unfairly plotted with just one force (i.e., much of the behavior is lost with just one "mean subpopulation force").
Therefore, plotting the forces between parent and offspring in objective space seems like a better possibility, although some aspects of the global movement is lost.

\begin{figure*}[h!]
 \centering
 \includegraphics[scale=0.8]{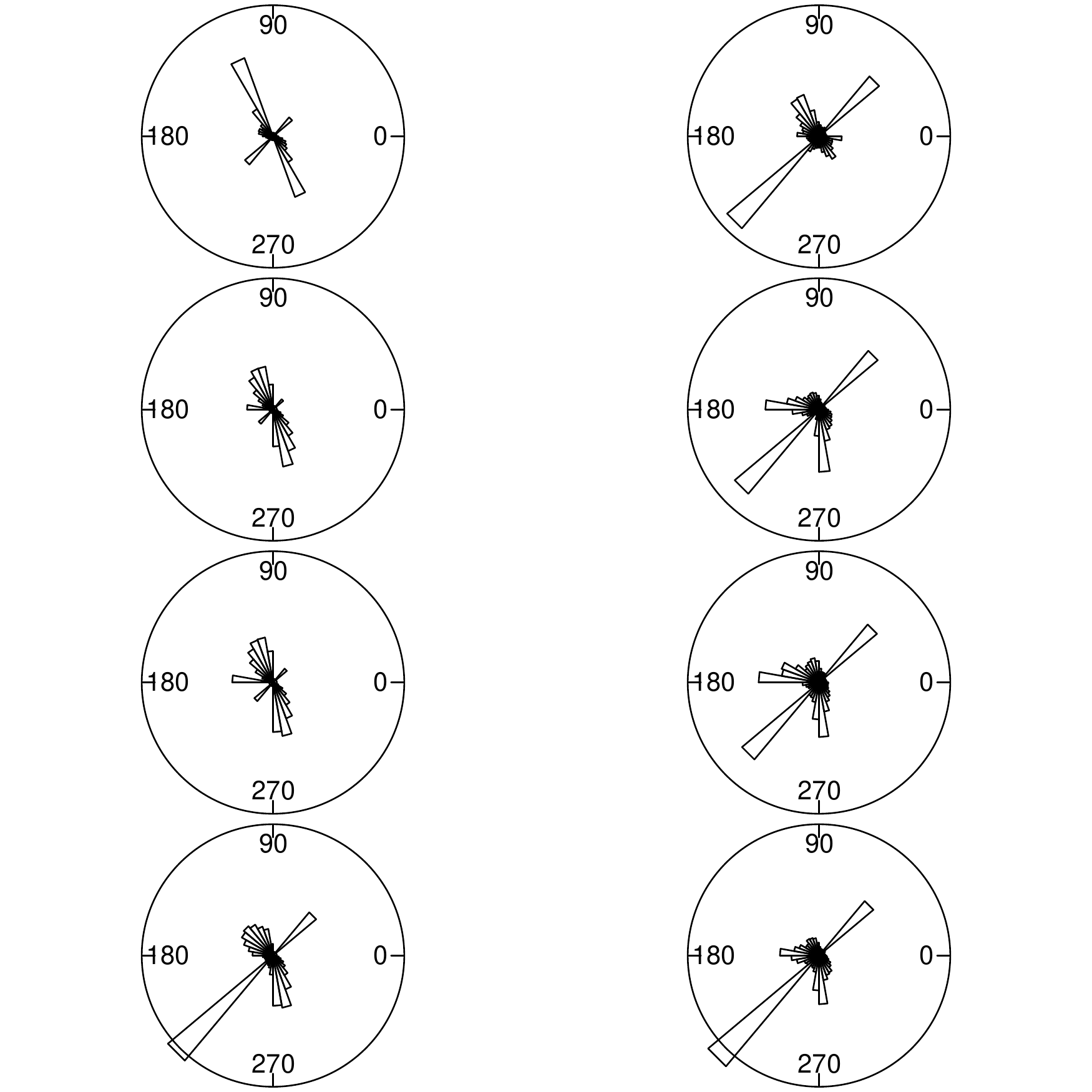}
 \caption{Accumulative angles of the forces measured for algorithms GDE3 (left column) and SAN (right column) on problems WFG1 (first row), WFG4 (second row), WFG5 (third row) and WFG8 (fourth row). The forces are measured by calculating the vector from the parent to the offspring in objective space. The scale is linear and the unfeasible solutions as well as forces with zero modulus were eliminated from the graph.}
 \label{rose}
\end{figure*}

Here, the forces are calculated by measuring the vector from the DE's operator main parent to its offspring in objective space (other genetic operators with no main parent might make necessary the computation of a set of forces for each individual with each force related to a parent).
The experiment is composed of $2500000$ evaluations samples throughout one run of the algorithm (multiple runs of the algorithm presented no significant difference from each other, as one would expect since the number of samples in one run are already representative).
Figure~\ref{rose} shows the accumulative angles of the forces for three problems with both GDE3 and SAN algorithms.
The direction given by the 0$^\circ$ and 90$^\circ$ are respectivelly parallel to increasing x-axis and increasing y-axis (i.e., 180$^\circ$  is improving objective 1, 270$^\circ$  is improving objective 2).
Regarding the measurement, it is done right before the selection phase of the differential evolution operator, otherwise the arc from 0$^\circ$ to 90$^\circ$ would be nonexistent. 
Naturally, a long bin means a higher number of solutions moving in that direction.
Notice, however, that some histograms may have more individuals than others. 
This happens because two conditions cause some solutions or forces to be discarded:
\begin{itemize}
	\item Unfeasible Solutions - They are excluded from the calculation, since unfeasible solutions can not be mapped to a point in objective space.
	\item Forces with Zero Modulus - In the case where the resulting child possess the same point in objective space as its main parent, the resulting force would have a zero modulus. In fact, this means that no force was applied at all and therefore it is reasonable to exclude it.
\end{itemize}

To give an idea of how many solutions were discarded and from which type (unfeasible solutions or solutions which result in a zero modulus force), Table~\ref{excluded_solutions} was constructed.
Setting problem WFG8 aside, GDE3 has always a high number of solutions discarded (specially solutions which result in a zero modulus force).
This happens because GDE3 converges prematuraly on these problems.
In WFG8, however, the solutions which result in a zero modulus force are extremely small.
This points to the fact that both algorithms have not yet converged in WFG8, explaining why GDE3 surpassed SAN in this problem.

\begin{table}
\centering
\caption{Percentage of solutions which are either unfeasible or result in zero modulus forces. Both types of solutions are excluded from the calculation of forces and therefore not present in Figure~\ref{rose}.}
\begin{tabular}{|l|c|c|c|c|}
\hline

Problem & \multicolumn{2}{c|}{Unfeasible Solutions} & \multicolumn{2}{c|}{Zero Modulus Forces} \\ \hline
	& SAN & GDE3 & SAN & GDE3\\ \hline
WFG1	& $15.64\%$ & $9.76\%$ & $13.38\%$  & $50.45\%$\\ \hline
WFG4	& $15.78\%$ & $3.81\%$ & $11.27\%$  & $55.47\%$\\ \hline
WFG5	& $27.44\%$ & $13.85\%$ & $8.11\%$  & $44.54\%$\\ \hline
WFG8	& $24.42\%$ & $11.06\%$ & $0.06\%$  & $0.34\%$\\ \hline
\end{tabular}
\label{excluded_solutions}
\end{table}

Bare in mind that the forces seen are not just a "DNA" of the algorithm.
They are affected intensively by the problem at hand.
Therefore, the higher the bias of the problem is, the higher the influence of the problem in the measured forces becomes.
The results on WFG1 and WFG8 shows exactly this interference of the problem which is strongly biased (see Table~\ref{wfg_properties} for the bias properties of all problems).
Therefore, analysing the behavior on problems with less bias (such as problems WFG4 and WFG5) renders a less noisy perspective on the "DNA" of the algorithm.
In fact, there are many similarities between the second and third rows of Figure~\ref{rose} with Figures~\ref{gde3_force} and~\ref{san_force}.
For example, the spread of forces in all directions can be seen in SAN, i.e., every direction has a bin with noticeable longness, while GDE3 has bins on fewer directions.
These results were predicted by our previous analysis.

The essential idea behind all these explanations is that a panmictic population may be seen as a niche.
Once no proper division is placed between strategies, no matter what strategies and procedures are involved, the population results in forces of different intensity being developed together, i.e., a conflict of forces appears inside the population.
This internal population conflict is hardly solved without a division.
That is, a division into subpopulations.

\subsection{Further Investigations}

The objective of this paper is to propose the framework together with some examples of algorithms based on it, demonstrating some of its aspects and strengths.
This is however not an exhaustive exposition. There are still an extensive amount of topics to be covered. To cite some:
\begin{itemize}
\item Studies on the variations of ${\cal IM}$ and ${\cal S}$ as well as self-adaptive modifications;
\item The effect of different and/or complex dynamics between subpopulations;
\item Integration of different types of algorithms and comparison between them.
\end{itemize}

\section{Conclusions}
\label{conclusion}

We have presented here a justification of why structured EAs, and in special the GSF, achieve better results in multi-objective optimization. 
This derives from the fact that \emph{well-designed structured EAs separate better the conflicting strategies, avoiding the deleterious consequences of the competition between themselves.}

Additionally, this article presented a new framework called GSF which can aid the understanding and design of structured optimization algorithms.
GSF can easily join any optimization algorithms, therefore any algorithm can be with little effort combined and tested together with others, yielding a very flexible framework.

Moreover, to the knowledge of the authors, SAN's results is the most or among the most robust algorithms of the state of the art, either surpassing GDE3 in the tests or achieving a comparable solution in terms of a trade-off between $\epsilon$ and hypervolume quality indicators.
In fact, when the problems increased in the number of objectives (which also increased the number of conflicting strategies inside a panmictic algorithm) the advantage of SAN over GDE3 became more emphatic. 
In other words, the proposed subpopulation framework showed that with an integration of simple algorithms it was possible to achieve better solutions, surpassing or at least achieving similar performance in all tests realized with the original panmictic algorithms.
Another interesting result is that a simple algorithm such as MONA, which had poor results on all tests, was shown to attain state of the art quality Pareto fronts when combined with two simple single-objective DEs in the subpopulation framework.

Thus, motivated by the population internal conflicts, structured optimization algorithms should find increasing attention of the optimization community.
In this aspect, the proposed subpopulation framework will hopefully aid the development of new structured algorithms and open new possibilities for the algorithms to come.
Consequently, further studies on multiple subpopulation dynamics as well as global interactions for the further understanding of the framework's frontiers is hereby encouraged.




%

\small

\bibliographystyle{apalike}
\bibliography{genetic_algorithm,article}

\end{document}